\documentclass[10pt,twocolumn,letterpaper]{article}

\usepackage{cvpr}
\usepackage{times}
\usepackage{epsfig}
\usepackage{graphicx}
\usepackage{amsmath}
\usepackage{amssymb}
\usepackage{booktabs}
\usepackage{multirow}

\usepackage[pagebackref=true,breaklinks=true,letterpaper=true,colorlinks,bookmarks=false]{hyperref}

\cvprfinalcopy 

\def\cvprPaperID{0129} 

\ifcvprfinal\fi
\begin{document}

\title{Mining Object Parts from CNNs via Active Question-Answering}

\author{Quanshi Zhang, Ruiming Cao, Ying Nian Wu, and Song-Chun Zhu\\
University of California, Los Angeles}

\maketitle

\begin{abstract}
Given a convolutional neural network (CNN) that is pre-trained for object classification, this paper proposes to use active question-answering to semanticize neural patterns in conv-layers of the CNN and mine part concepts. For each part concept, we mine neural patterns in the pre-trained CNN, which are related to the target part, and use these patterns to construct an And-Or graph (AOG) to represent a four-layer semantic hierarchy of the part. As an interpretable model, the AOG associates different CNN units with different explicit object parts. We use an active human-computer communication to incrementally grow such an AOG on the pre-trained CNN as follows. We allow the computer to actively identify objects, whose neural patterns cannot be explained by the current AOG. Then, the computer asks human about the unexplained objects, and uses the answers to automatically discover certain CNN patterns corresponding to the missing knowledge. We incrementally grow the  AOG to encode new knowledge discovered during the active-learning process. In experiments, our method exhibits high learning efficiency. Our method uses about $1/6$--$1/3$ of the part annotations for training, but achieves similar or better part-localization performance than fast-RCNN methods.
\end{abstract}

\section{Introduction}

\begin{figure}[t]
\centering
\includegraphics[width=0.99\linewidth]{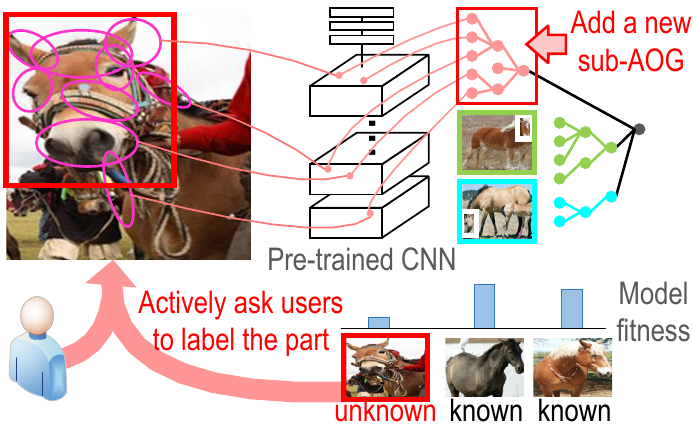}
\caption{Semanticizing knowledge in a pre-trained CNN via active question-answering (QA). We mine latent patterns from the CNN to explain certain object parts, and organize such patterns into a semantic hierarchy. Our method automatically identifies objects whose parts cannot be explained by part templates in the current AOG, asks about the objects, and uses the answers to mine patterns from these objects. The mined patterns represent new part templates, and are organized as new branches in the AOG.}
\label{fig:top}
\vspace{-15pt}
\end{figure}

Convolutional neural networks (CNNs)~\cite{CNN,CNNImageNet} have been trained to achieve near human-level performance on object detection. However, CNN methods still face two issues in real-world applications. First, many visual tasks require detailed interpretations of object structures for hierarchical understanding of objects (\emph{e.g.} part localization and parsing). This is beyond the detection of object bounding boxes. Second, weakly-supervised learning is also a difficult problem for CNNs. Unlike data-rich applications (\emph{e.g.} pedestrian/vehicle detection), many tasks require modeling certain object parts on the fly. For example, people may hope to use only a few examples to quickly teach a robot how to grasp a certain type of object parts for an occasional task.

In this study, we propose a new strategy to model a certain object part using a few part annotations, \emph{i.e.} \textit{using an active question-answering (QA) process to mine latent patterns that are related to the part from a pre-trained CNN. We use an And-Or graph (AOG) as an interpretable model to associate these patterns with the target part.}

We develop our method based on the following three  ideas: \textbf{1)} When a CNN is pre-trained using objects of a category with object-box annotations, most appearance knowledge of the target category may have been encoded in conv-layers of the CNN. \textbf{2)} Our task is to mine latent patterns from complex neural activations in the conv-layers. Each pattern individually acts as a detector of a certain region of an object. We use the mined regional patterns to construct an AOG to represent the target part. \textbf{3)} Because the AOG represents the part's neural patterns with clear semantic hierarchy, we can start an active QA to incrementally grow new AOG branches to encode new part templates, so as to enrich the knowledge in the AOG.

More specifically, during the active QA, the computer discovers objects whose neural activations cannot be explained by the current AOG and asks human users for supervision. We use the answers to grow new AOG branches for new part templates given in answers. Active QA makes the part knowledge efficiently learned with very limited human supervision.

\textbf{CNN generalization:}{\verb| |} Before we introduce inputs and outputs of our QA-based learning, we clarify our target of CNN generalization, \emph{i.e.} growing semantic AOGs to explain semantic hierarchy hidden within the conv-layers of a pre-trained CNN.

\begin{figure}[t]
\centering
\includegraphics[width=0.99\linewidth]{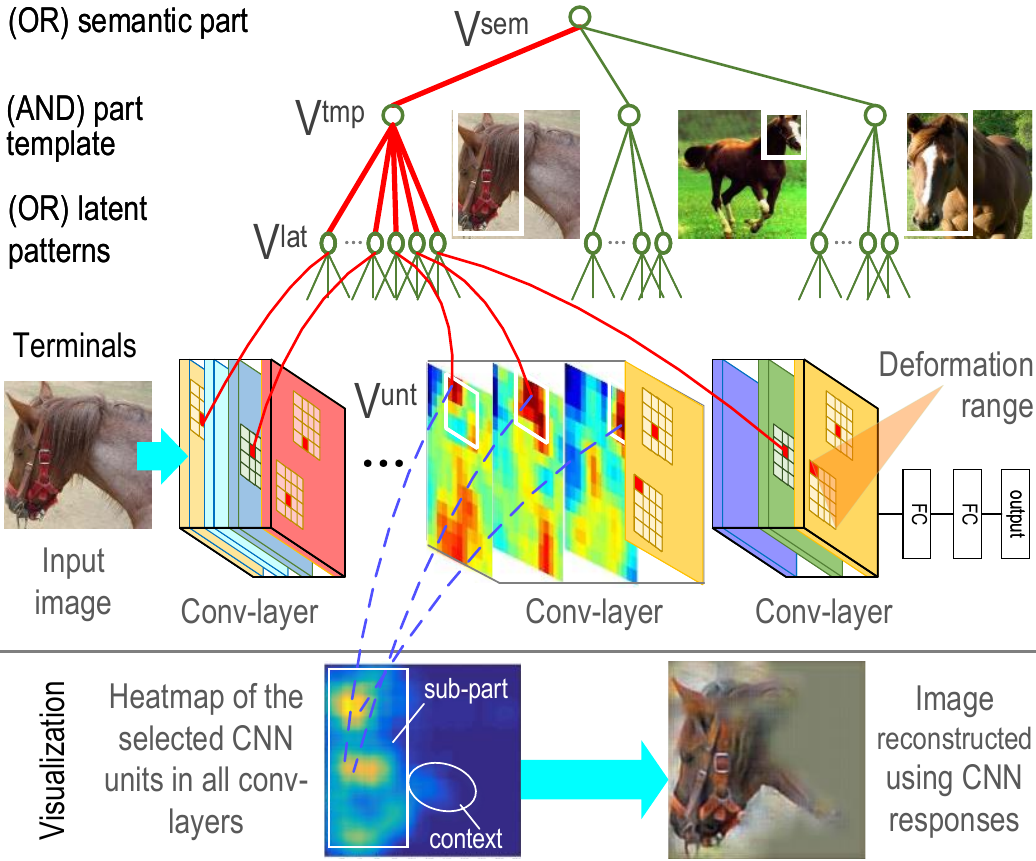}
\caption{And-Or graph grown on the pre-trained CNN as a semantic branch. The AOG associates certain CNN units with certain image regions. The red lines indicate the parse graph.}
\label{fig:model}
\vspace{-15pt}
\end{figure}

As shown in Fig~\ref{fig:model}, the AOG has four layers, which encode a clear semantic hierarchy ranging from \textit{semantic part}, \textit{part templates}, \textit{latent patterns}, to \textit{CNN units}. In the AOG, we use AND nodes to represent compositional regions of a part, and use OR nodes to encode a list of alternative template/deformation candidates for a local region. The top \textit{part} node (OR node) uses its children to represent a number of template candidates for the part. Each \textit{part template} (AND node) in the second layer has a number of children as latent patterns to represent its constituent regions (\emph{e.g.} an eye in the face part). Each \textit{latent pattern} in the third layer (OR node) naturally corresponds to a certain range of units within a CNN conv-slice. We select a \textit{CNN unit} within this range to account for geometric deformation of the latent pattern.

Note that we do not further fine-tune the original convolutional weights within the pre-trained CNN. This allows us to continuously grow AOGs for different parts, without the risk of model drifting.

\textbf{Inputs and outputs of QA-based learning:}{\verb| |} Given a pre-trained CNN and its training samples (\emph{i.e.} object images without any part annotations), we incrementally grow AOG branches for the target part. In each step of QA, we let the CNN use the current AOG to localize the target part among all the unannotated images. Our method actively identifies object images, whose parts cannot be well explained by the AOG. Among all the unexplained objects, our method predicts the potential gain of asking about each unexplained object, and thus determines a best sequence of questions for QA. As in Fig.~\ref{fig:QA}, the user is able to give five types of answers to explicitly guide the AOG growth. Given each specific answer, the computer may refine an existing part template or mine latent patterns to construct a new AOG branch for a new part template.

\textbf{Learning from weak supervision:}{\verb| |} Unlike previous end-to-end batch learning, there are two mechanisms to ensure the stability of weakly-supervised learning. 1) We transfer patterns in a pre-trained object-level CNN to the target part concept, instead of learning all knowledge from scratch. These patterns are supposed to consistently describe the same part region among different object images. The pattern-mining process purifies the CNN knowledge for better representation of the target part. 2) We use active QA to collect training samples, in order to avoid wasting human labor of annotating object parts that can be well explained by the AOG.

We use object-level annotations for pre-training, considering the following two facts: 1) Only a few datasets~\cite{SemanticPart,CUB200} provide part annotations, and most benchmark datasets~\cite{PascalVOC,ImageNet,MSCOCO} mainly have annotations of object bounding boxes. 2) More crucially, different applications may focus on different object parts, and it is impractical to annotate a large number of parts for each specific task.

\textbf{Contributions:}{\verb| |} Contributions of this study can be summarized as follows. 1) We mine and represent latent patterns hidden in a pre-trained CNN using an AOG. The AOG representation enables the QA \emph{w.r.t} the semantic hierarchy of the target part. 2) We propose to use active QA to explicitly learn the semantics of each AOG branch, which ensures a high learning efficiency. 3) In experiments, our method exhibits superior performance to other baselines in terms of weakly-supervised part localization. For example, our methods with 11 part annotations outperformed fast-RCNNs with 60 annotations in Fig.~\ref{fig:curve}.

\section{Related work}

\textbf{Passive CNN visualization vs. active CNN semanticization:}{\verb| |} In order to explore the hidden semantics in the CNN, many studies visualized and analyzed patterns of CNN units~\cite{CNNVisualization_1,CNNVisualization_2,CNNVisualization_3,CNNVisualization_5,CNNFeatureMining}.

However, from the perspective of semanticizing CNN units, CNN visualization and our active QA go in two opposite directions. Given a certain unit in a pre-trained CNN, the former mainly visualizes the potential visual pattern of the unit passively. However, the latter focuses on a more fundamental problem in real applications, \emph{i.e.} \textit{given a query of modeling/refining certain object parts, can we efficiently discover certain patterns that are related to the part concepts, within the pre-trained CNN from its complex neural activations?} Given CNN feature maps, Zhou~\emph{et al.}~\cite{CNNSemanticDeep,CNNSemanticDeep2} discovered latent ``scene'' semantics. Simon~\emph{et al.} discovered objects~\cite{ObjectDiscoveryCNN_2} from CNN activations in an unsupervised manner, and learned part concepts in a supervised fashion~\cite{CNNSemanticPart}. AOG structure is suitable for representing semantic hierarchy of objects~\cite{MumfordAOG,MiningAOG}, and \cite{CNNAoG} used an AOG to represent the CNN. In this study, we used semantic-level QA to incrementally mine part semantics from the CNN and grow the AOG. Such a ``white-box'' representation of the CNN knowledge also guided further active QA.

\textbf{Unsupervised/active learning:}{\verb| |} Many methods have been developed to learn object models in an unsupervised or weakly supervised manner. Methods of \cite{Gpt_WeaklyCNN,WeaklyMIL,OurICCV15AoG,DiscoveryCNNFeature} learned with image-level annotations without labeling object bounding boxes. \cite{UnsuperCNN,ChoDiscovery} did not require any annotations during the learning process. \cite{OnlineMetric}  collected training data online from videos to incrementally learn models. \cite{Language2VideoAlign,Language2ActionAlign} discovered objects and identified actions from language Instructions and videos. Inspired by active learning~\cite{Active4,i13,Active2}, the idea of learning from question-answering has been used to learn object models~\cite{KB_Fei_Annotation,KB_Fei_InteractionLabel,TuQA}. Branson \emph{et al.}~\cite{ActivePart} used human-computer interactions to label object parts to learn part models. Instead of directly building new models from active QA, our method uses the QA to semanticize the CNN and transfer the hidden knowledge to the AOG.


\textbf{Modeling ``objects'' vs. modeling ``parts'' in un-/weakly-supervised learning:}{\verb| |} In the scope of unsupervised learning and/or weakly-supervised learning, modeling parts is usually more challenging than modeling entire objects. Given image-level labels (without object bounding boxes), object discovery~\cite{ObjectDiscoveryCNN_1,ObjectDiscoveryCNN_2,ObjectDiscoveryCNN_3} and co-segmentation~\cite{InteractiveCoseg} can be achieved by identifying common foreground patterns from complex background. In addition, there are some strong prior knowledges for object discovery, such as closed boundaries and common object structures.

In contrast, to the best of our knowledge, there is no mechanism to distinguish a certain part concept from other parts of the same object. It is because 1) all the parts represent common foreground patterns among objects; 2) some parts (\emph{e.g.} the abdomen) do not have shape boundaries to identify their shape extent. Thus, up to now, people mainly extract implicit middle-level part patches~\cite{MiddleLevel}, but it is difficult to capture explicit semantic meanings of these parts.

\section{Preliminaries: And-Or graph on a CNN}

In this section, we briefly introduce an AOG, which is designed to explain the latent semantic structure within the CNN. As shown in Fig.~\ref{fig:model}, an AOG has four layers, \emph{i.e.} \textit{semantic part} (OR node), \textit{part template} (AND node), \textit{latent pattern} (OR node), and \textit{CNN unit}. In the AOG, an OR node encodes a number of alternative candidates as children. An AND node uses its children to represent its constituent regions. For example, 1) the semantic part (OR node) encodes a number of template candidates for the part as children. 2) Each part template (AND node) encodes the spatial relationship between its children latent patterns (each child corresponds to a constituent region or a contextual image region). 3) Each latent pattern (OR node) takes a number of CNN units in a certain conv-slice as children to represent alternative deformation candidates of the pattern (the pattern may appear in different image positions).

Given an image $I$\textcolor{red}{\footnote[1]{Considering CNN's superior performance in object detection, as in \cite{SemanticPart}, we regard object detection and part localization as two separate processes for evaluation. Thus, we crop $I$ to only contain the object and resize $I$ for CNN inputs to simplify the scenario of learning for part localization.}}, we use the CNN to compute neural activations on $I$ in its conv-layers, and then use the AOG for \textit{hierarchical part parsing}. \emph{I.e.} we use the AOG to semanticize the neural activations and localize the target part.

We use $V^{\textrm{sem}}$, $V^{\textrm{tmp}}\!\in\!\Omega^{\textrm{tmp}}$, $V^{\textrm{lat}}\!\in\!\Omega^{\textrm{lat}}$, and $V^{\textrm{unt}}\!\in\!\Omega^{\textrm{unt}}$, respectively, to denote nodes at the four layers. During the parsing procedure, 1) the top node $V^{\textrm{sem}}$ selects a part template {\small$\hat{V}^{\textrm{tmp}}$} to explain the whole part; 2) {\small$\hat{V}^{\textrm{tmp}}$} let its children latent patterns use their own parsing configurations to vote for {\small$\hat{V}^{\textrm{tmp}}$}'s position, thereby parsing an image region for {\small$\hat{V}^{\textrm{tmp}}$}; 3) each latent pattern {\small$V^{\textrm{lat}}\in Child(\hat{V}^{\textrm{tmp}})$} selects a CNN-unit child with a certain deformation range {\small$\hat{V}^{\textrm{unt}}\in Child(V^{\textrm{lat}})$} as a stand-in of the pattern.

We define a parse graph $pg_{I}$ to denote the parsing configurations. As the red lines in Fig.~\ref{fig:model}, $pg_{I}$ is a tree of image regions that are assigned to AOG nodes, {\small$pg\!=\!\{\Lambda_{I,V^{\textrm{sem}}}\}\cup\{\Lambda_{I,\hat{V}^{\textrm{tmp}}}\}\cup_{V^{\textrm{lat}}\in Child(\hat{V}^{\textrm{tmp}})}\{\Lambda_{I,V^{\textrm{lat}}}\}$}, where for each node $V$, $\Lambda_{I,V}$ denotes the image region that is parsed for $V$. We use $\Lambda_{V}$ to simplify the notation of $\Lambda_{I,V}$, without ambiguity.

We design an inference score $S_{I}(V|\Lambda_{V})$ for each node $V$ to measure the compatibility between a given region $\Lambda_{V}$ and $V$ (as well as the AOG branch under $V$). Thus, hierarchical part parsing on a given image $I$ can be achieved in a bottom-up manner. We compute inference scores for CNN units, then propagate the scores to latent patterns and part templates, and finally obtain the score of the top node as the overall inference score {\small${\bf L}(I,{\boldsymbol\theta})$}. We determine the parse graph $\hat{pg}_{I}$ that maximizes the overall score:
\begin{small}
\vspace{-3pt}
\begin{equation}
\!\!\!{\bf L}(I,{\boldsymbol\theta})\!=\!S_{I}(V^{\textrm{sem}}|\Lambda_{V^{\textrm{sem}}}),\quad\hat{pg}_{I}\!=\!{\arg\!\max}_{pg_{I}}{\bf L}(I,{\boldsymbol\theta})|_{pg_{I}}\!\!\!
\label{eqn:parsing}
\vspace{-3pt}
\end{equation}
\end{small}
where ${\boldsymbol\theta}$ denotes the AOG parameters.

\textbf{Terminal nodes (CNN units):} Each terminal node under a latent pattern takes a certain square within a certain conv-slice, which represents deformation candidates of the latent pattern. Each $V^{\textrm{unt}}$ corresponds to a fixed image region {\small$\Lambda_{V^{\textrm{unt}}}$}. \emph{I.e.} we propagate $V^{\textrm{unt}}$'s receptive field to the image plane, and use the final field as {\small$\Lambda_{V^{\textrm{unt}}}$}. The score of {\small$V^{\textrm{unt}}$}, {\small$S_{I}(V^{\textrm{unt}})$}\textcolor{red}{\footnote[2]{Please see \cite{CNNAoG} for detailed settings.}}, is designed to describe the neural response value of {\small$V^{\textrm{unt}}$} and its local deformation level.

\textbf{OR nodes:} Given children's parsing configurations of an OR node (either {\small$V^{\textrm{sem}}$} or {\small$V^{\textrm{lat}}$}), {\small$V^{O}$} selects the child {\small$\hat{V}$} with the highest score, and propagates {\small$\hat{V}$}'s parsing result to {\small$V^{O}$}:
\begin{small}
\vspace{-3pt}
\begin{equation}
\!\!\!S_{I}(V^{O}|\hat{\Lambda}_{V^{O}})\!=\!{\max}_{V\in Child(V^{O})}S_{I}(V|\hat{\Lambda}_{V}),\quad\hat{\Lambda}_{V^{O}}\!\leftarrow\!\hat{\Lambda}_{\hat{V}}\!\!\!
\vspace{-10pt}
\end{equation}
\end{small}

\textbf{AND nodes:} Given parsing results of a part template $V^{\textrm{tmp}}$'s children latent patterns, we parse an image region for $V^{\textrm{tmp}}$, which maximizes its score.
\begin{small}
\vspace{-3pt}
\begin{equation}
\begin{split}
\!\!\!&\!\!S_{I}(V^{\textrm{tmp}}\!|\Lambda_{V^{\textrm{tmp}}})\!=\!\!\!\!\!\!\!\!\!\!\!\!\!\!\sum_{V^{\textrm{lat}}\in\!Child(V^{\textrm{tmp}})}\!\!\!\!\!\!\!\!\!\!\!\!\big[S_{I}(V^{\textrm{lat}}|\hat{\Lambda}_{V^{\textrm{lat}}})+S^{\textrm{inf}}(\Lambda_{V^{\textrm{tmp}}}|\hat{\Lambda}_{V^{\textrm{lat}}})\big]\!\!\!\!\!\!\!\!\!\!\!\\
\!&\hat{\Lambda}_{V^{\textrm{tmp}}}={\arg\!\max}_{\Lambda_{V^{\textrm{tmp}}}}S_{I}(V^{\textrm{tmp}}|\Lambda_{V^{\textrm{tmp}}})
\end{split}
\vspace{-5pt}
\end{equation}
\end{small}
where $S^{\textrm{inf}}(\Lambda_{V^{\textrm{tmp}}}|\hat{\Lambda}_{V^{\textrm{lat}}})$\textcolor{red}{\footnotemark[2]} measures the spatial compatibility between parsing configurations of $\Lambda_{V^{\textrm{tmp}}}$ and $\hat{\Lambda}_{V^{\textrm{lat}}}$ on $I$.



\textbf{AOG construction:}{\verb| |} The method for constructing an AOG based on part annotations was proposed in \cite{CNNAoG}. We briefly summarize this method as follows. Let ${\bf I}$ denote a set of cropped object images of a category. Among all objects in ${\bf I}$, only a small number of objects, ${\bf I}^{\textrm{ant}}\!=\!\{I_{i}|i\!=\!1,2,\ldots,m\}\subset{\bf I}$, have annotations of the target part. For each annotated object {\small$I\in{\bf I}^{\textrm{ant}}$}, we label two terms {\small$(\Lambda_{I,V^{\textrm{sem}}}^{*},V_{I}^{\textrm{tmp}*})$}. {\small$\Lambda_{I,V_{I}^{\textrm{sem}}}^{*}$} denotes the ground-truth bounding box of the part, and $V_{I}^{\textrm{tmp}*}$ specifies the true choice of the part template for the part in $I$. For the first two layers of the AOG, the AOG is set to only contain the part templates that appear in part annotations.

Thus, AOG construction is to mine a total of $n$ different latent patterns for each part template {\small$V^{\textrm{tmp}}$}, where $n$ is a hyper-parameter. For each latent pattern {\small$V^{\textrm{lat}}$}, parameters {\small${\boldsymbol\theta}_{V^{\textrm{lat}}}\!\subset\!{\boldsymbol\theta}$} mainly determine 1) {\small$V^{\textrm{lat}}$}'s deformation range and 2) the prior displacement from {\small$V^{\textrm{tmp}}$} to {\small$V^{\textrm{lat}}$}. The estimation of {\small${\boldsymbol\theta}_{V^{\textrm{lat}}}$} can be roughly written as\textcolor{red}{\footnotemark[2]}
\begin{small}
\vspace{-3pt}
\begin{equation}
\!\!\!\!\underset{\boldsymbol\theta}{\max}\Big\{\underset{I\in{\bf I}_{V^{\textrm{tmp}}}}{\textrm{mean}}\,S_{I}(V_{I}^{\textrm{tmp}*}|\Lambda_{V_{I}^{\textrm{tmp}*}}\!=\!\Lambda_{I,V^{\textrm{sem}}}^{*})+\underset{I'\in{\bf I}}{\textrm{mean}}\,S^{\textrm{local}}_{I}(V^{\textrm{tmp}})\Big\}\!\!\!
\label{eqn:build}
\vspace{-5pt}
\end{equation}
\end{small}
where {\small${\bf I}_{V^{\textrm{tmp}}}\!=\!\{I\in{\bf I}^{\textrm{ant}}|V_{I}^{\textrm{tmp}*}\!=\!V^{\textrm{tmp}}\}$}. Compared to {\small$S_{I}(V_{I}^{\textrm{tmp}*}|\Lambda_{V_{I}^{\textrm{tmp}*}})$},
{\small$S^{\textrm{local}}_{I}(V^{\textrm{tmp}})$ $=\sum_{V^{\textrm{lat}}\in\!Child(V^{\textrm{tmp}})}$ $S_{I}(V^{\textrm{lat}}$ $|\hat{\Lambda}_{V^{\textrm{lat}}})$} is an inference score that ignores the pairwise spatial compatibility.


\begin{figure}[t]
\centering
\includegraphics[width=0.99\linewidth]{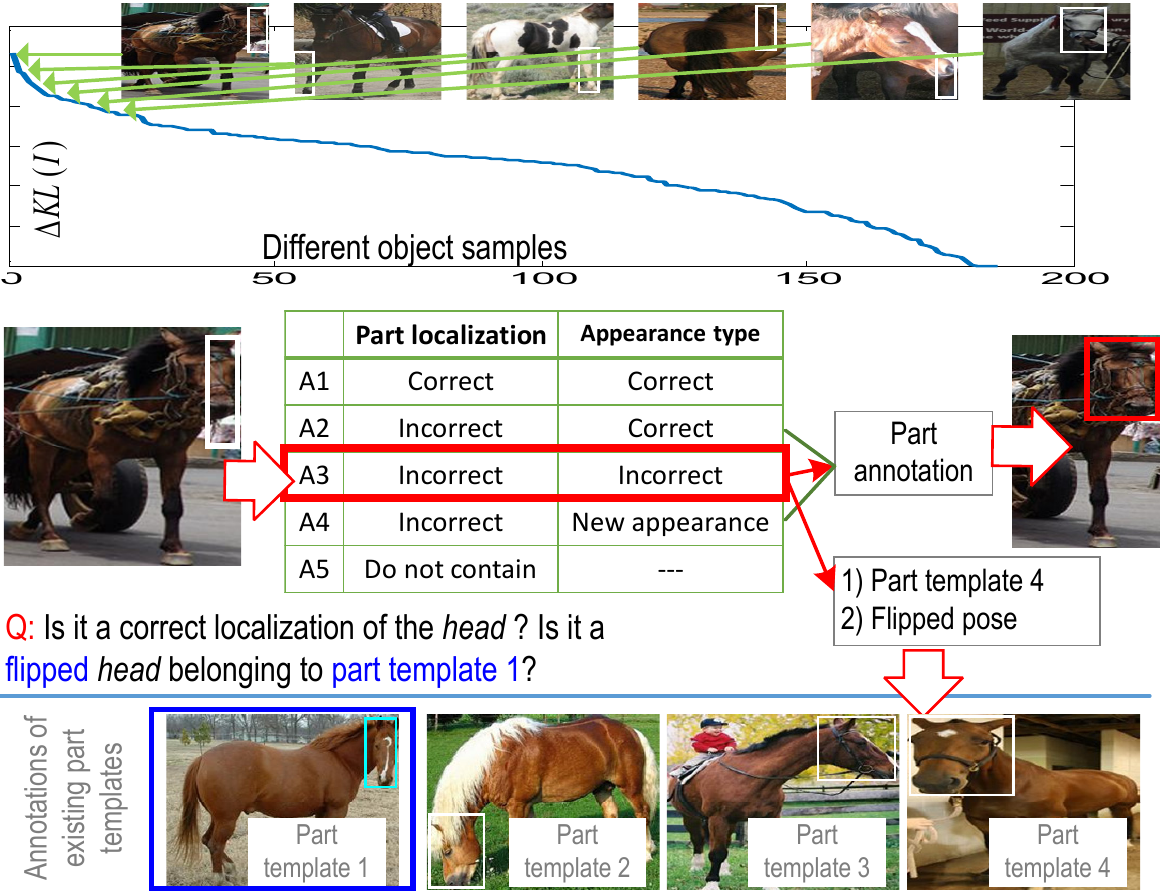}
\caption{Illustration of the QA process. (top) We sort and select objects. (bottom) We show questions asked for each target object.}
\label{fig:QA}
\vspace{-15pt}
\end{figure}

\section{Learning from active question-answering}

\subsection{Overview of knowledge mining}

Compared to conventional batch learning, our method uses a more efficient learning strategy, which allows the computer to actively detect blind spots in its knowledge system and ask questions. In general, knowledge blind spots in the AOG include 1) neural-activation patterns in the CNN that have not been modeled and 2) the inaccuracy of the existing latent patterns. We assume that the unexplained neural patterns potentially reflect new part templates, while the inaccurate latent patterns correspond to the sub-optimally modeled part templates.

Because an AOG is an interpretable representation that explicitly encodes object parts, we can represent blind spots of the knowledge using linguistic description. We use a total of five types of answers to explicitly project these blind spots onto specific semantic details of objects. In this way, the computer selects and asks a series of questions. Based on the answers, the AOG incrementally grows new semantic branches to explain new part templates and refine AOG branches of existing part templates.

The computer repeats the following process in each QA step. Let ${\bf I}$ denote a set of object images. As shown in Fig.~\ref{fig:QA}, the computer first uses the current AOG to localize object parts on all unannotated objects in ${\bf I}$. Based on localization results, the computer selects and asks about an object $I$, from which the computer believes it can obtain the most information gain. A question {\small$q\!=\!(I,\hat{V}^{\textrm{tmp}},\hat{\Lambda}_{V^{\textrm{sem}}})$} requires people to determine whether the computer determines the correct part template $\hat{V}^{\textrm{tmp}}$ and accurately localizes the part in $\hat{\Lambda}_{V^{\textrm{sem}}}$, and expects one of the following answers.

\textbf{Answer 1:} the part detection is correct. \textbf{Answer 2:} the computer chooses the true template for the part in the parse graph, but it does not accurately localizes the target part. \textbf{Answer 3:} neither the part template nor the part location is correctly estimated. \textbf{Answer 4:} the part belongs to a new part template. \textbf{Answer 5:} the target part does not appear in the object. In addition, in case of receiving Answers~2--4, the computer will ask people to annotate the target part. In case of getting Answer 3, the computer will require people to specify the part template, as well as whether the object is flipped. Then, our method uses the new annotation to refine (for Answers 2--3) or create (for Answer 4) the AOG branch for the annotated part template based on Eq.~(\ref{eqn:build}).

\subsection{Question ranking}

The core of the QA process is to select a sequence of objects that reduce the AOG uncertainty the most. Therefore, in this section, we design a loss function to measure the incompatibility between the AOG knowledge and the actual part appearance in the object samples. We predict the potential gain (decrease of the loss) of asking about each object. Objects with large gains usually correspond to unexplained or not well explained CNN neural activations. Note that annotating the part in an object may also help explain parts on other objects, thereby leading to a large gain. Thus, we use a greedy strategy to select a sequence of questions {\small$\Omega=\{q_{i}|i=1,2,\ldots\}$}, \emph{i.e.} asking about the object that leads to the most gain in each step.

For each object $I\in{\bf I}$, we use ${\bf P}(y|I)$ and ${\bf Q}(y|I)$ to denote the prior distribution and the estimated distribution of an object part on $I$, respectively. $y\in\{+1,-1\}$ is a label indicating whether $I$ contains the target part. The current AOG estimates the probability of object $I$ containing the target part as {\small${\bf Q}(y\!=\!+1|I)\!=\!\frac{1}{Z}\exp[\beta{\bf L}(I,{\boldsymbol\theta})]$}, where $Z$ and $\beta$ are scaling parameters (see Section~\ref{sec:implement} for details); {\small${\bf Q}(y\!=\!-1|I)\!=\!1-{\bf Q}(y\!=\!+1|I)$}. Let {\small${\bf I}^{\textrm{ask}}\subset{\bf I}$} denotes the objects that have been asked during previous QA. For each asked object {\small$I\in{\bf I}^{\textrm{ask}}$}, we set its prior distribution {\small${\bf P}(y\!=\!+1|I)\!=\!1$} if $I$ contains the target part according to previous answers; {\small${\bf P}(y=\!+1\!|I)\!=\!0$} otherwise. For each un-asked object {\small$I\in{\bf I}\setminus{\bf I}^{\textrm{ask}}$}, we set its prior distribution based on statistics of previous answers, {\small${\bf P}(y\!=\!+1|I)\!=\!\textrm{mean}_{I'\in{\bf I}^{\textrm{ask}}}{\bf P}(y\!=\!+1|I')$}. Therefore, we formulate the loss function as the KL divergence between the prior distribution ${\bf P}$ and the estimated distribution ${\bf Q}$, and seek to minimize the KL divergence via QA.
\begin{small}
\vspace{-3pt}
\begin{equation}
\begin{split}
\!\!\!Loss\!=\!{\bf KL}({\bf P}\Vert{\bf Q})=&{\sum}_{I\in{\bf I}}{\sum}_{y}{\bf P}(y,I)\log\frac{{\bf P}(y,I)}{{\bf Q}(y,I)}\!\!\!\!\!\!\!\\
=&\lambda{\sum}_{I\in{\bf I}}{\sum}_{y}{\bf P}(y|I)\log\frac{{\bf P}(y|I)}{{\bf Q}(y|I)}
\end{split}
\vspace{-5pt}
\end{equation}
\end{small}
where {\small${\bf P}(y,I)\!=\!{\bf P}(y|I)P(I)$; ${\bf Q}(y,I)\!=\!{\bf Q}(y|I)P(I)$; $\lambda=P(I)\!=\!1/\vert{\bf I}\vert$} is a constant prior probability for object $I$.

In fact, both the prior distribution ${\bf P}$ and the estimated distribution ${\bf Q}$ keep changing during the QA process. Let us assume that the computer selects object {\small$\tilde{I}\in{\bf I}\setminus{\bf I}^{\textrm{ask}}$} and that people annotate its part. The annotation would encode the part knowledge of $\tilde{I}$ into the AOG and greatly change the estimated distribution for objects that are similar to $\tilde{I}$. For each object {\small$I'\in{\bf I}$}, we predict its estimated distribution after the new part annotation as
\begin{small}
\vspace{-3pt}
\begin{equation}
\begin{split}
\tilde{\bf Q}(y\!=\!+1|I')=&\frac{1}{Z}\exp[\beta{\bf L}(I',{\boldsymbol\theta}^{\textrm{new}})|_{\tilde{I}}]\\
{\bf L}(I',{\boldsymbol\theta}^{\textrm{new}})|_{\tilde{I}}=&{\bf L}(I',{\boldsymbol\theta})+\Delta{\bf L}(\tilde{I},{\boldsymbol\theta})e^{-\alpha\cdot dist(I',\tilde{I})}\!\!\!\!\!\!\!\!\!\!
\end{split}
\label{eqn:predict}
\vspace{-5pt}
\end{equation}
\end{small}
where ${\bf L}(I',{\boldsymbol\theta}^{\textrm{new}})|_{\tilde{I}}$ indicates the predicted inference score of $I'$ when we annotate $\tilde{I}$. We assume that if object $I'$ is similar to object $\tilde{I}$, the inference score of $I'$ will have an increase similar to that of $\tilde{I}$. We estimate the score increase of $\tilde{I}$ as {\small$\Delta{\bf L}(\tilde{I},{\boldsymbol\theta})\!=\!\textrm{mean}_{I\in{\bf I}^{\textrm{ant}}}{\bf L}(I,{\boldsymbol\theta})-{\bf L}(\tilde{I},{\boldsymbol\theta})$}. $\alpha$ is a scalar weight. We formulate the appearance distance between $I'$ and $\tilde{I}$ as {\small$dist(I',\tilde{I})\!=\!1-\frac{\phi(I')^{T}\phi(\tilde{I})}{\vert\phi(I')\vert\cdot\vert\phi(\tilde{I})\vert}$}, where {\small$\phi(I')\!=\!{\bf M}\,{\bf f}_{I'}$}. ${\bf f}_{I'}$ denotes CNN features of $I'$ at the top conv-layer after ReLu operation, and ${\bf M}$ is a diagonal matrix representing the prior reliability for each feature dimension\textcolor{red}{\footnote[3]{${\bf M}_{ii}\!\propto\!\exp[{\textrm{mean}}_{I\in{\bf I}}S_{I}(V^{\textrm{unt}}_{i})]$, where $V^{\textrm{unt}}_{i}$ is the CNN unit corresponding to the $i$-th element of ${\bf f}_{I'}$.}}. Thus, {\small$\exp[\alpha\cdot dist(I',\tilde{I})]$} measures the similarity between $I'$ and $\tilde{I}$. In addition, if $I'$ and $\tilde{I}$ are assigned with different part templates by the current AOG, we may ignore the similarity between $I'$ and $\tilde{I}$ (by setting an infinite distance between them) to achieve better performance. Based on the prediction in Eq.~(\ref{eqn:predict}), we can predict the changes of the KL divergence after the new annotation on $\tilde{I}$ as
\begin{small}
\vspace{-3pt}
\begin{equation}
\Delta{\bf KL}(\tilde{I})=\lambda{\sum}_{I\in{\bf I}}{\sum}_{y}{\bf P}(y|I)\log\frac{\tilde{\bf Q}(y|I)}{{\bf Q}(y|I)}
\label{eqn:delta}
\vspace{-5pt}
\end{equation}
\end{small}
Thus, in each step, the computer selects and asks about the object that maximize the decrease of the KL divergence.
\begin{small}
\vspace{-3pt}
\begin{equation}
\hat{I}={\arg\!\max}_{I\in{\bf I}\setminus{\bf I}^{\textrm{ask}}}\Delta{\bf KL}(I)
\label{eqn:select}
\vspace{-10pt}
\end{equation}
\end{small}

\textbf{QA implementations:}{\verb| |} In the beginning, for each object $I$, we initialize its prior distribution as ${\bf P}(y\!=\!+1|I)\!=\!1$ and its estimated distribution as ${\bf Q}(y\!=\!+1|I)\!=\!0$. Then, the computer selects and asks about an object $\hat{I}$ based on Eq.~(\ref{eqn:select}). We use the answer to update ${\bf P}$. If new object parts are labeled during the QA process, we apply Eq.~(\ref{eqn:build}) to update the AOG. More specifically, if people label a new part template, the AOG will grow a new AOG branch to encode this template. If people annotate a part for an old part template, our method will update its corresponding AOG branch. Then, the new AOG can provide the new distribution  ${\bf Q}$. In later steps, the computer repeats the above QA procedure of Eq.~(\ref{eqn:select}) and Eq.~(\ref{eqn:build}) to ask more questions.

\section{Experiments}

\subsection{Implementation details}
\label{sec:implement}

We used the 16-layer VGG network (VGG-16)~\cite{VGG}, which was pre-trained using 1.3M images in the ImageNet ILSVRC 2012 dataset~\cite{ImageNet} with a loss for 1000-category classification. Then, in order to learn part concepts for each category, we further fine-tune the VGG-16 using object images in this category based on the loss for classifying target objects and background. The VGG-16 contains a total of 13 conv-layers and 3 fully connected layers. We selected the last 9 conv-layers as valid conv-layers. We extracted CNN units from these layers to build the AOG.

In our method, three parameters were involved in active QA, \emph{i.e.} $\alpha$, $\beta$, and $Z$. Considering that most object images contained the target part in real applications, we ignored the small probability of ${\bf P}(y\!=\!-1|I)$ in Eq.~(\ref{eqn:delta}) to simplify the computation. As a result, the parameter $Z$ was eliminated in the computation of Eq.~(\ref{eqn:delta}), and the parameter $\beta$ acted as a constant weight for $\Delta{\bf KL}(\tilde{I})$, which did not affect object selection in Eq.~(\ref{eqn:select}). Therefore, in our experiments, we set $\alpha\!=\!4.0$, which achieved the best performance.

\subsection{Datasets}

We used three benchmark datasets to test our method, \emph{i.e.} the PASCAL VOC Part Dataset~\cite{SemanticPart}, the CUB200-2011 dataset~\cite{CUB200}, and the ILSVRC 2013 DET Animal-Part dataset~\cite{CNNAoG}. Just like in most part-localization studies~\cite{SemanticPart,CNNAoG}, we selected animal categories, which prevalently contain non-rigid shape deformation, to test part-localization performance. \emph{I.e.} we selected six animal categories---\textit{bird, cat, cow, dog, horse}, and \textit{sheep}---from the PASCAL Part Dataset. The CUB200-2011 dataset contains 11.8K images of 200 bird species. Like in \cite{ActivePart,CNNSemanticPart,CNNAoG}, we ignored species labels and regarded all these images as a single bird category. The ILSVRC 2013 DET Animal-Part dataset~\cite{CNNAoG} was proposed for part localization. It consists of 30 animal categories among all the 200 categories for object detection in the ILSVRC 2013 DET dataset~\cite{ImageNet}.

\begin{table}[t]
\centering\begin{small}
\resizebox{0.75\linewidth}{!}{\begin{tabular}{c|ccc}
\hline
\!\!\!Annotation\!\!\!&\!\!\! \;\;\;Layer 1:\!\!\!&\!\!\! \;\;\;Layer 2:\!\!\!&\!\!\! \;\;\;Layer 3:\!\!\!\\
\!\!\!number\!\!\!&\!\!\!semantic part\!\!\!&\!\!\!part template\!\!\!&\!\!\!latent pattern\!\!\!\\
\hline
\!\!\!{05}
\!\!\!&\!\!\! {3.15} \!\!\!&\!\!\! {3791.5} \!\!\!&\!\!\! {91.6} \!\!\!\\
\hline
\!\!\!{10}
\!\!\!&\!\!\! {5.95} \!\!\!&\!\!\! {3804.8} \!\!\!&\!\!\! {93.9} \!\!\!\\
\hline
\!\!\!{15}
\!\!\!&\!\!\! {8.52} \!\!\!&\!\!\! {3760.4} \!\!\!&\!\!\! {95.5} \!\!\!\\
\hline
\!\!\!{20}
\!\!\!&\!\!\! {11.16} \!\!\!&\!\!\! {3778.3} \!\!\!&\!\!\! {96.3} \!\!\!\\
\hline
\!\!\!{25}
\!\!\!&\!\!\! {13.55} \!\!\!&\!\!\! {3777.5} \!\!\!&\!\!\! {98.3} \!\!\!\\
\hline
\!\!\!{30}
\!\!\!&\!\!\! {15.83} \!\!\!&\!\!\! {3837.3} \!\!\!&\!\!\! {99.2} \!\!\!\\
\hline
\end{tabular}}
\caption{Average number of children of AOG nodes}
\label{tab:stat}
\vspace{-15pt}
\end{small}
\end{table}

\begin{table*}[t]
\centering
\resizebox{1.0\linewidth}{!}{\begin{tabular}{p{3.3cm}|c|c|cccccccccccccccc}
\hline
\!\!\!&\!\! Part Annot. \!\!\!&\!\! \!{\footnotesize Obj.-box finetune}\! \!\!\!&\!\! gold. \!\!\!&\!\! bird \!\!\!&\!\! frog \!\!\!&\!\! turt. \!\!\!&\!\! liza. \!\!\!&\!\! koala \!\!\!&\!\! lobs. \!\!\!&\!\! dog \!\!\!&\!\! fox \!\!\!&\!\! cat \!\!\!&\!\! lion \!\!\!&\!\! tiger \!\!\!&\!\! bear \!\!\!&\!\! rabb. \!\!\!&\!\! hams. \!\!\!&\!\! squi.\\
\!\!\! SS-DPM-Part~\cite{SSDPM} \!\!\!&\!\! 60 \!\!\!&\!\!  No
\!\!\!&\!\!0.1859
\!\!\!&\!\!0.2747
\!\!\!&\!\!0.2105
\!\!\!&\!\!0.2316
\!\!\!&\!\!0.2901
\!\!\!&\!\!0.1755
\!\!\!&\!\!0.1666
\!\!\!&\!\!0.1948
\!\!\!&\!\!0.1845
\!\!\!&\!\!0.1944
\!\!\!&\!\!0.1334
\!\!\!&\!\!0.0929
\!\!\!&\!\!0.1981
\!\!\!&\!\!0.1355
\!\!\!&\!\!0.1137
\!\!\!&\!\!0.1717
\\
\!\!\! PL-DPM-Part~\cite{PLDPM} \!\!\!&\!\! 60  \!\!\!&\!\! No
\!\!\!&\!\!0.2867
\!\!\!&\!\!0.2337
\!\!\!&\!\!0.2169
\!\!\!&\!\!0.2650
\!\!\!&\!\!0.3079
\!\!\!&\!\!0.1445
\!\!\!&\!\!0.1526
\!\!\!&\!\!0.1904
\!\!\!&\!\!0.2252
\!\!\!&\!\!0.1488
\!\!\!&\!\!0.1450
\!\!\!&\!\!0.1340
\!\!\!&\!\!0.1838
\!\!\!&\!\!0.1968
\!\!\!&\!\!0.1389
\!\!\!&\!\!0.2590
\\
\!\!\! Part-Graph~\cite{SemanticPart} \!\!\!&\!\! 60 \!\!\!&\!\! No
\!\!\!&\!\!0.3385
\!\!\!&\!\!0.3305
\!\!\!&\!\!0.3853
\!\!\!&\!\!0.2873
\!\!\!&\!\!0.3813
\!\!\!&\!\!0.0848
\!\!\!&\!\!0.3467
\!\!\!&\!\!0.1679
\!\!\!&\!\!0.1736
\!\!\!&\!\!0.3499
\!\!\!&\!\!0.1551
\!\!\!&\!\!0.1225
\!\!\!&\!\!0.1906
\!\!\!&\!\!0.2068
\!\!\!&\!\!0.1622
\!\!\!&\!\!0.3038
\\
\!\!\! fc7+linearSVM \!\!\!&\!\! 60 \!\!\!&\!\! Yes
\!\!\!&\!\!0.1359
\!\!\!&\!\!0.2117
\!\!\!&\!\!0.1681
\!\!\!&\!\!0.1890
\!\!\!&\!\!0.2557
\!\!\!&\!\!0.1734
\!\!\!&\!\!0.1845
\!\!\!&\!\!0.1451
\!\!\!&\!\!0.1374
\!\!\!&\!\!0.1581
\!\!\!&\!\!0.1528
\!\!\!&\!\!0.1525
\!\!\!&\!\!0.1354
\!\!\!&\!\!0.1478
\!\!\!&\!\!0.1287
\!\!\!&\!\!0.1291
\\
\!\!\! fc7+RBF-SVM \!\!\!&\!\! 60 \!\!\!&\!\! Yes
\!\!\!&\!\!0.1818
\!\!\!&\!\!0.2637
\!\!\!&\!\!0.2035
\!\!\!&\!\!0.2246
\!\!\!&\!\!0.2538
\!\!\!&\!\!0.1663
\!\!\!&\!\!0.1660
\!\!\!&\!\!0.1512
\!\!\!&\!\!0.1670
\!\!\!&\!\!0.1719
\!\!\!&\!\!0.1176
\!\!\!&\!\!0.1638
\!\!\!&\!\!0.1325
\!\!\!&\!\!0.1312
\!\!\!&\!\!0.1410
\!\!\!&\!\!0.1343
\\
\!\!\! CNN-PDD~\cite{CNNSemanticPart} \!\!\!&\!\! 60 \!\!\!&\!\! No
\!\!\!&\!\!0.1932
\!\!\!&\!\!0.2015
\!\!\!&\!\!0.2734
\!\!\!&\!\!0.2195
\!\!\!&\!\!0.2650
\!\!\!&\!\!0.1432
\!\!\!&\!\!0.1535
\!\!\!&\!\!0.1657
\!\!\!&\!\!0.1510
\!\!\!&\!\!0.1787
\!\!\!&\!\!0.1560
\!\!\!&\!\!0.1756
\!\!\!&\!\!0.1444
\!\!\!&\!\!0.1320
\!\!\!&\!\!0.1251
\!\!\!&\!\!0.1776
\\
\!\!\! CNN-PDD-ft~\cite{CNNSemanticPart} \!\!\!&\!\! 60 \!\!\!&\!\! Yes
\!\!\!&\!\!0.2109
\!\!\!&\!\!0.2531
\!\!\!&\!\!0.1999
\!\!\!&\!\!0.2144
\!\!\!&\!\!0.2494
\!\!\!&\!\!0.1577
\!\!\!&\!\!0.1605
\!\!\!&\!\!0.1847
\!\!\!&\!\!0.1845
\!\!\!&\!\!0.2127
\!\!\!&\!\!0.1521
\!\!\!&\!\!0.2066
\!\!\!&\!\!0.1826
\!\!\!&\!\!0.1595
\!\!\!&\!\!0.1570
\!\!\!&\!\!0.1608
\\
\!\!\! Fast-RCNN (1 ft)~\cite{FastRCNN} \!\!\!&\!\! 30 \!\!\!&\!\! No
\!\!\!&\!\!0.0847
\!\!\!&\!\!0.1520
\!\!\!&\!\!0.1905
\!\!\!&\!\!0.1696
\!\!\!&\!\!0.1412
\!\!\!&\!\!0.0754
\!\!\!&\!\!0.2538
\!\!\!&\!\!0.1471
\!\!\!&\!\!0.0886
\!\!\!&\!\!0.0944
\!\!\!&\!\!0.1004
\!\!\!&\!\!0.0585
\!\!\!&\!\!0.1013
\!\!\!&\!\!0.0821
\!\!\!&\!\!0.0577
\!\!\!&\!\!0.1005
\\
\!\!\! Fast-RCNN (2 fts)~\cite{FastRCNN} \!\!\!&\!\! 30 \!\!\!&\!\! Yes
\!\!\!&\!\!0.0913
\!\!\!&\!\!0.1043
\!\!\!&\!\!0.1294
\!\!\!&\!\!0.1632
\!\!\!&\!\!0.1585
\!\!\!&\!\!0.0730
\!\!\!&\!\!0.2530
\!\!\!&\!\!0.1148
\!\!\!&\!\!0.0736
\!\!\!&\!\!{\bf0.0770}
\!\!\!&\!\!0.0680
\!\!\!&\!\!{\bf0.0441}
\!\!\!&\!\!0.1265
\!\!\!&\!\!0.1017
\!\!\!&\!\!0.0709
\!\!\!&\!\!0.0834
\\
\!\!\! Ours \!\!\!&\!\! \textcolor{red}{\bf 10} \!\!\!&\!\! Yes
\!\!\!&\!\!{\bf 0.0796}
\!\!\!&\!\!{\bf 0.0850}
\!\!\!&\!\!{\bf 0.0906}
\!\!\!&\!\!0.2077
\!\!\!&\!\!{\bf 0.1260}
\!\!\!&\!\!0.0759
\!\!\!&\!\!{\bf 0.1212}
\!\!\!&\!\!0.1476
\!\!\!&\!\!{\bf 0.0584}
\!\!\!&\!\!0.1107
\!\!\!&\!\!0.0716
\!\!\!&\!\!0.0637
\!\!\!&\!\!0.1092
\!\!\!&\!\!{\bf 0.0755}
\!\!\!&\!\!0.0697
\!\!\!&\!\!{\bf 0.0421}
\\
\!\!\! Ours \!\!\!&\!\! \textcolor{red}{\bf 20} \!\!\!&\!\! Yes
\!\!\!&\!\!{\bf 0.0638}
\!\!\!&\!\!{\bf 0.0793}
\!\!\!&\!\!{\bf 0.0765}
\!\!\!&\!\!{\bf 0.1221}
\!\!\!&\!\!{\bf 0.1174}
\!\!\!&\!\!{\bf 0.0720}
\!\!\!&\!\!{\bf 0.1201}
\!\!\!&\!\!{\bf 0.1096}
\!\!\!&\!\!{\bf 0.0517}
\!\!\!&\!\!0.1006
\!\!\!&\!\!0.0752
\!\!\!&\!\!0.0624
\!\!\!&\!\!0.1090
\!\!\!&\!\!{\bf 0.0788}
\!\!\!&\!\!0.0603
\!\!\!&\!\!{\bf 0.0454}
\\
\!\!\! Ours \!\!\!&\!\! \textcolor{red}{\bf 30} \!\!\!&\!\! Yes
\!\!\!&\!\!{\bf 0.0642}
\!\!\!&\!\!{\bf 0.0734}
\!\!\!&\!\!{\bf 0.0971}
\!\!\!&\!\!{\bf 0.0916}
\!\!\!&\!\!{\bf 0.0948}
\!\!\!&\!\!{\bf 0.0658}
\!\!\!&\!\!{\bf 0.1355}
\!\!\!&\!\!{\bf 0.1023}
\!\!\!&\!\!{\bf 0.0474}
\!\!\!&\!\!0.1011
\!\!\!&\!\!{\bf 0.0625}
\!\!\!&\!\!0.0632
\!\!\!&\!\!{\bf 0.0964}
\!\!\!&\!\!{\bf 0.0783}
\!\!\!&\!\!{\bf 0.0540}
\!\!\!&\!\!{\bf 0.0499}
\\
\!\!\!&\!\! \!\!\!&\!\! \!\!\!&\!\! horse \!\!\!&\!\! zebra \!\!\!&\!\! swine \!\!\!&\!\! hippo \!\!\!&\!\! catt. \!\!\!&\!\! sheep \!\!\!&\!\! ante. \!\!\!&\!\! camel \!\!\!&\!\! otter \!\!\!&\!\! arma. \!\!\!&\!\! monk. \!\!\!&\!\! elep. \!\!\!&\!\! red pa. \!\!\!&\!\! gia.pa. \!\!\!&\!\! \!\!\!&\!\! \textcolor{blue}{\bf\large Avg.}\\
\!\!\! SS-DPM-Part~\cite{SSDPM} \!\!\!&\!\! 60 \!\!\!&\!\! No
\!\!\!&\!\!0.2346
\!\!\!&\!\!0.1717
\!\!\!&\!\!0.2262
\!\!\!&\!\!0.2261
\!\!\!&\!\!0.2371
\!\!\!&\!\!0.2364
\!\!\!&\!\!0.2026
\!\!\!&\!\!0.2308
\!\!\!&\!\!0.2088
\!\!\!&\!\!0.2881
\!\!\!&\!\!0.1859
\!\!\!&\!\!0.1740
\!\!\!&\!\!0.1619
\!\!\!&\!\!0.0989
\!\!\!&\!\!
\!\!\!&\!\!\textcolor{blue}{0.1946}
\\
\!\!\! PL-DPM-Part~\cite{PLDPM} \!\!\!&\!\! 60 \!\!\!&\!\! No
\!\!\!&\!\!0.2657
\!\!\!&\!\!0.2937
\!\!\!&\!\!0.2164
\!\!\!&\!\!0.2150
\!\!\!&\!\!0.2320
\!\!\!&\!\!0.2145
\!\!\!&\!\!0.3119
\!\!\!&\!\!0.2949
\!\!\!&\!\!0.2468
\!\!\!&\!\!0.3100
\!\!\!&\!\!0.2113
\!\!\!&\!\!0.1975
\!\!\!&\!\!0.1835
\!\!\!&\!\!0.1396
\!\!\!&\!\!
\!\!\!&\!\!\textcolor{blue}{0.2187}
\\
\!\!\! Part-Graph~\cite{SemanticPart} \!\!\!&\!\! 60 \!\!\!&\!\! No
\!\!\!&\!\!0.2804
\!\!\!&\!\!0.3376
\!\!\!&\!\!0.2979
\!\!\!&\!\!0.2964
\!\!\!&\!\!0.2513
\!\!\!&\!\!0.2321
\!\!\!&\!\!0.3504
\!\!\!&\!\!0.2179
\!\!\!&\!\!0.2535
\!\!\!&\!\!0.2778
\!\!\!&\!\!0.2321
\!\!\!&\!\!0.1961
\!\!\!&\!\!0.1713
\!\!\!&\!\!0.0759
\!\!\!&\!\!
\!\!\!&\!\!\textcolor{blue}{0.2486}
\\
\!\!\! fc7+linearSVM \!\!\!&\!\! 60 \!\!\!&\!\! Yes
\!\!\!&\!\!0.2003
\!\!\!&\!\!0.2409
\!\!\!&\!\!0.1632
\!\!\!&\!\!0.1400
\!\!\!&\!\!0.2043
\!\!\!&\!\!0.2274
\!\!\!&\!\!0.1479
\!\!\!&\!\!0.2204
\!\!\!&\!\!0.2498
\!\!\!&\!\!0.2875
\!\!\!&\!\!0.2261
\!\!\!&\!\!0.1520
\!\!\!&\!\!0.1557
\!\!\!&\!\!0.1071
\!\!\!&\!\!
\!\!\!&\!\!\textcolor{blue}{0.1776}
\\
\!\!\! fc7+RBF-SVM \!\!\!&\!\! 60 \!\!\!&\!\! Yes
\!\!\!&\!\!0.2207
\!\!\!&\!\!0.1550
\!\!\!&\!\!0.1963
\!\!\!&\!\!0.1536
\!\!\!&\!\!0.2609
\!\!\!&\!\!0.2295
\!\!\!&\!\!0.1748
\!\!\!&\!\!0.2080
\!\!\!&\!\!0.2263
\!\!\!&\!\!0.2613
\!\!\!&\!\!0.2244
\!\!\!&\!\!0.1806
\!\!\!&\!\!0.1417
\!\!\!&\!\!0.1095
\!\!\!&\!\!
\!\!\!&\!\!\textcolor{blue}{0.1838}
\\
\!\!\! CNN-PDD~\cite{CNNSemanticPart} \!\!\!&\!\! 60 \!\!\!&\!\! No
\!\!\!&\!\!0.2610
\!\!\!&\!\!0.2363
\!\!\!&\!\!0.1623
\!\!\!&\!\!0.2018
\!\!\!&\!\!0.1955
\!\!\!&\!\!0.1350
\!\!\!&\!\!0.1857
\!\!\!&\!\!0.2499
\!\!\!&\!\!0.2486
\!\!\!&\!\!0.2656
\!\!\!&\!\!0.1704
\!\!\!&\!\!0.1765
\!\!\!&\!\!0.1713
\!\!\!&\!\!0.1638
\!\!\!&\!\!
\!\!\!&\!\!\textcolor{blue}{0.1893}
\\
\!\!\! CNN-PDD-ft~\cite{CNNSemanticPart} \!\!\!&\!\! 60 \!\!\!&\!\! Yes
\!\!\!&\!\!0.2417
\!\!\!&\!\!0.2725
\!\!\!&\!\!0.1943
\!\!\!&\!\!0.2299
\!\!\!&\!\!0.2104
\!\!\!&\!\!0.1936
\!\!\!&\!\!0.1712
\!\!\!&\!\!0.2552
\!\!\!&\!\!0.2110
\!\!\!&\!\!0.2726
\!\!\!&\!\!0.1463
\!\!\!&\!\!0.1602
\!\!\!&\!\!0.1868
\!\!\!&\!\!0.1475
\!\!\!&\!\!
\!\!\!&\!\!\textcolor{blue}{0.1980}
\\
\!\!\! Fast-RCNN (1 ft)~\cite{FastRCNN} \!\!\!&\!\! 30 \!\!\!&\!\! No
\!\!\!&\!\!0.2694
\!\!\!&\!\!{\bf0.0823}
\!\!\!&\!\!0.1319
\!\!\!&\!\!0.0976
\!\!\!&\!\!0.1309
\!\!\!&\!\!0.1276
\!\!\!&\!\!0.1348
\!\!\!&\!\!0.1609
\!\!\!&\!\!0.1627
\!\!\!&\!\!0.1889
\!\!\!&\!\!0.1367
\!\!\!&\!\!{\bf0.1081}
\!\!\!&\!\!0.0791
\!\!\!&\!\!{\bf0.0474}
\!\!\!&\!\!
\!\!\!&\!\!\textcolor{blue}{0.1252}
\\
\!\!\! Fast-RCNN (2 fts)~\cite{FastRCNN} \!\!\!&\!\! 30 \!\!\!&\!\! Yes
\!\!\!&\!\!0.1629
\!\!\!&\!\!0.0881
\!\!\!&\!\!{\bf0.1228}
\!\!\!&\!\!{\bf0.0889}
\!\!\!&\!\!{\bf0.0922}
\!\!\!&\!\!{\bf0.0622}
\!\!\!&\!\!0.1000
\!\!\!&\!\!0.1519
\!\!\!&\!\!0.0969
\!\!\!&\!\!{\bf0.1485}
\!\!\!&\!\!0.0855
\!\!\!&\!\!0.1085
\!\!\!&\!\!{\bf0.0407}
\!\!\!&\!\!0.0542
\!\!\!&\!\!
\!\!\!&\!\!\textcolor{blue}{0.1045}
\\
\!\!\! Ours \!\!\!&\!\! \textcolor{red}{\bf 10} \!\!\!&\!\! Yes
\!\!\!&\!\!{\bf 0.1297}
\!\!\!&\!\!0.1413
\!\!\!&\!\!0.2145
\!\!\!&\!\!0.1377
\!\!\!&\!\!0.1493
\!\!\!&\!\!0.1415
\!\!\!&\!\!0.1046
\!\!\!&\!\!{\bf 0.1239}
\!\!\!&\!\!0.1288
\!\!\!&\!\!0.1964
\!\!\!&\!\!{\bf 0.0524}
\!\!\!&\!\!0.1507
\!\!\!&\!\!0.1081
\!\!\!&\!\!0.0640
\!\!\!&\!\!
\!\!\!&\!\!\textcolor{blue}{0.1126}
\\
\!\!\! Ours \!\!\!&\!\! \textcolor{red}{\bf 20} \!\!\!&\!\! Yes
\!\!\!&\!\!{\bf 0.1083}
\!\!\!&\!\!0.1389
\!\!\!&\!\!0.1475
\!\!\!&\!\!0.1280
\!\!\!&\!\!0.1490
\!\!\!&\!\!0.1300
\!\!\!&\!\!{\bf 0.0667}
\!\!\!&\!\!{\bf 0.1033}
\!\!\!&\!\!0.1103
\!\!\!&\!\!0.1526
\!\!\!&\!\!{\bf 0.0497}
\!\!\!&\!\!0.1301
\!\!\!&\!\!0.0802
\!\!\!&\!\!0.0574
\!\!\!&\!\!
\!\!\!&\!\!\textcolor{blue}{\bf 0.0965}
\\
\!\!\! Ours \!\!\!&\!\! \textcolor{red}{\bf 30} \!\!\!&\!\! Yes
\!\!\!&\!\!{\bf 0.1129}
\!\!\!&\!\!0.1066
\!\!\!&\!\!0.1408
\!\!\!&\!\!0.1204
\!\!\!&\!\!0.1118
\!\!\!&\!\!0.1260
\!\!\!&\!\!{\bf 0.0825}
\!\!\!&\!\!{\bf 0.0836}
\!\!\!&\!\!{\bf 0.0901}
\!\!\!&\!\!0.1685
\!\!\!&\!\!{\bf 0.0490}
\!\!\!&\!\!0.1224
\!\!\!&\!\!0.0779
\!\!\!&\!\!0.0577
\!\!\!&\!\!
\!\!\!&\!\!\textcolor{blue}{\bf 0.0909}
\\
\hline
\end{tabular}}
\caption{Normalized distance of part localization on the ILSVRC 2013 DET Animal-Part dataset. The second column shows the number of part annotations for training. The third column indicates whether the baseline used all object-box annotations in the category to pre-fine-tune a CNN before learning the part (\textit{object-box annotations are more than part annotations}).}
\label{tab:imgnet}
\vspace{-12pt}
\end{table*}

\subsection{Baselines}

We compared the proposed method with the following thirteen baselines. We designed the first two baselines based on the Fast-RCNN~\cite{FastRCNN}. Note that we fine-tuned the fast-RCNN with a loss for detecting a single class/part from background, rather than for multi-class/part detection, for a fair comparison. In the first baseline, namely \textit{Fast-RCNN (1 ft)}, we directly fine-tuned the VGG-16 network using part annotations to detect parts on well cropped objects. Then, to enable a fair comparison, we conducted the second baseline based on two-stage fine-tuning, namely \textit{Fast-RCNN (2 fts)}. \textit{The Fast-RCNN (2 fts) first fine-tuned the VGG-16 network using a large number of object-box annotations {\rm(more than part annotations)} in the target category, and then fine-tuned the VGG-16 using a few part annotations.}

The third baseline was proposed by \cite{CNNSemanticPart}, namely \textit{CNN-PDD}. \textit{CNN-PDD} selected a conv-slice in a CNN (pre-trained using ImageNet ILSVRC 2012 dataset) to represent and localize the part on well cropped objects. Then, we slightly extended \cite{CNNSemanticPart} as the fourth baseline \textit{CNN-PDD-ft}. \textit{CNN-PDD-ft} fine-tuned the VGG-16 using object-box annotations, and then applied \cite{CNNSemanticPart} to the VGG-16 for learning.

The fifth and sixth baselines were the strongly supervised DPM (\textit{SS-DPM-Part})~\cite{SSDPM} and the technique in \cite{PLDPM} (\textit{PL-DPM-Part}), respectively. They trained DPMs using part annotations for part localization. We used the graphical model proposed in \cite{SemanticPart} as the seventh baseline, namely \textit{Part-Graph}. The eighth baseline was the interactive learning of DPMs for part localization~\cite{ActivePart} (\textit{Interactive-DPM}).

Without many training samples, ``simple'' methods are usually insensitive to the over-fitting problem. Thus, we designed the last four baselines as follows. We used the VGG-16 network that was fine-tuned using object-box annotations, and collected image patches from a cropped object based on the selective search~\cite{SelectiveSearch}. We used the VGG-16 to extract \textit{fc7} features from each image patch. The two baselines (\emph{i.e.} \textit{fc7+linearSVM} and \textit{fc7+RBF-SVM}) used a linear SVM and a RBF-SVM, respectively, to detect the target part. The other baselines \textit{VAE+linearSVM} and \textit{CoopNet+linearSVM} used features of the VAE network~\cite{VAE} and the CoopNet~\cite{CoopNet}, respectively, instead of \textit{fc7} features, for part detection.

Finally, the last baseline is the learning of AOGs~\cite{CNNAoG} without QA (\textit{AOG w/o QA}). We annotated parts and part templates on randomly selected objects.

In fact, both object annotations and part annotations are used to learn models in all the thirteen baselines (including those without fine-tuning).


\subsection{Evaluation metric}

It has been discussed in \cite{SemanticPart,CNNAoG} that a fair evaluation of part localization requires removing the factors of object detection. Therefore, we used ground-truth object bounding boxes to crop objects from the original images to produce testing images. Given an object image, object/part detection methods (\emph{e.g.} \textit{Fast-RCNN (1 ft)}, \textit{Part-Graph}, and \textit{SS-DPM-Part}) usually estimate several bounding boxes for the part with different confidence values. As in \cite{CNNSemanticPart,SemanticPart,ObjectDiscoveryCNN_1,CNNAoG}, the task of part localization takes the most confident bounding box per image as the result. Given part-localization results on objects of a category, we applied the \textit{normalized distance}~\cite{CNNSemanticPart} and the \textit{percentage of correctly localized parts} (PCP)~\cite{fineGrained1,fineGrained2,fineGrained3} to evaluate part localization. For the normalized distance, we computed the distance between the predicted part center and the ground-truth part center, and then normalized the distance using the diagonal length of the object as the normalized distance. For PCP, we used the typical metric of ``$IoU\geq0.5$''~\cite{FastRCNN} to identify correct part localizations.

\begin{table}[t]
\centering
\resizebox{1.0\linewidth}{!}{\begin{tabular}{lcccc}
\hline
\!\!\!&\!\! \!\!\!\!\!\!\!\!\!\!\!\!{\small Obj.-box finetune} \!\!\!&\!\! {\small Part Annot.} \!\!\!&\!\! \!\#Q \!\!\!&\!\!{\small Normalizaed distance}\!\!\!\\
\!\!\! SS-DPM-Part~\cite{SSDPM} \!\!\!&\!\! No \!\!\!&\!\! 60 \!\!\!&\!\! -- \!\!\!& 0.2504\\
\!\!\! PL-DPM-Part~\cite{PLDPM} \!\!\!&\!\! No \!\!\!&\!\! 60 \!\!\!&\!\! -- \!\!\!& 0.3215\\
\!\!\! Part-Graph~\cite{SemanticPart} \!\!\!&\!\! No \!\!\!&\!\! 60 \!\!\!&\!\! -- \!\!\!& 0.3697\\
\!\!\! fc7+linearSVM \!\!\!&\!\! Yes \!\!\!&\!\! 60 \!\!\!&\!\! -- \!\!\!& 0.2786\\
\!\!\! fc7+RBF-SVM \!\!\!&\!\! Yes \!\!\!&\!\! 60 \!\!\!&\!\! -- \!\!\!& 0.3360\\
\!\!\! Interactive-DPM~\cite{ActivePart} \!\!\!&\!\! No \!\!\!&\!\! 60 \!\!\!&\!\! -- \!\!\! & 0.2011\\
\!\!\! CNN-PDD~\cite{CNNSemanticPart} \!\!\!&\!\! No \!\!\!&\!\! 60 \!\!\!&\!\! -- \!\!\! & 0.2446\\
\!\!\! CNN-PDD-ft~\cite{CNNSemanticPart} \!\!\!&\!\! Yes \!\!\!&\!\! 60 \!\!\!&\!\! -- \!\!\! & 0.2694\\
\!\!\! Fast-RCNN (1 ft)~\cite{FastRCNN} \!\!\!&\!\! No \!\!\!&\!\! 60 \!\!\!&\!\! -- \!\!\! & 0.3105\\
\!\!\! Fast-RCNN (2 fts)~\cite{FastRCNN} \!\!\!&\!\! Yes \!\!\!&\!\! 60 \!\!\!&\!\! -- \!\!\! & 0.1989\\
\!\!\! AOG w/o QA~\cite{CNNAoG} \!\!\!&\!\! Yes \!\!\!&\!\! \textcolor{red}{\bf 20} \!\!\!&\!\! -- \!\!\! & 0.1084\\
\!\!\! Ours \!\!\!&\!\! Yes \!\!\!&\!\! \textcolor{red}{\bf 10} \!\!\!&\!\! 28 \!\!\! & {\bf 0.0626}\\
\!\!\! Ours \!\!\!&\!\! Yes \!\!\!&\!\! \textcolor{red}{\bf 20} \!\!\!&\!\! 112 \!\!\! & {\bf 0.0434}\\
\hline
\end{tabular}}
\caption{Part localization performance on the CUB200-2011 dataset. See Table~\ref{tab:imgnet} for the introduction of the 2nd and 3rd columns. The 4rd column shows the number of questions for training. The fourth column indicates whether the baseline used all object annotations (\textit{more than part annotations}) in the category to pre-fine-tune a CNN before learning the part.}
\label{tab:cub200}
\vspace{-12pt}
\end{table}

\begin{table}[t]
\centering
\resizebox{1.0\linewidth}{!}{\begin{tabular}{c|lccccccccc}
\hline
\!\!\! & \!\!\! Method \!\!\!&\!\! \!\!\!\!\!\!Annot.\!\! \!\!\!&\!\! \#Q \!\!\!&\!\! bird \!\!\!&\!\! cat \!\!\!&\!\! cow \!\!\!&\!\! dog \!\!\!&\!\! {\small horse} \!\!\!&\!\! {\small sheep} \!\!\!&\!\! \textcolor{blue}{\bf Avg.}\!\!\!\\
\hline
\multirow{7}{*}{\rotatebox[origin=c]{90}{Head\vspace{-3pt}}} &\!\!\! {\small Fast-RCNN (1 ft)~\cite{FastRCNN}} \!\!\!&\!\! \textcolor{red}{\bf 10} \!\!\!&\!\! --
\!\!\!&\!\!0.326
\!\!\!&\!\!0.238
\!\!\!&\!\!0.283
\!\!\!&\!\!0.286
\!\!\!&\!\!0.319
\!\!\!&\!\!0.354
\!\!\!&\!\!\textcolor{blue}{0.301}\!\!\!
\\
& \!\!\! {\small Fast-RCNN (2 fts)~\cite{FastRCNN}} \!\!\!&\!\! \textcolor{red}{\bf 10} \!\!\!&\!\! --
\!\!\!&\!\!0.233
\!\!\!&\!\!0.196
\!\!\!&\!\!0.216
\!\!\!&\!\!0.206
\!\!\!&\!\!0.253
\!\!\!&\!\!0.286
\!\!\!&\!\!\textcolor{blue}{0.232}\!\!\!
\\
& \!\!\! {\small Fast-RCNN (1 ft)~\cite{FastRCNN}} \!\!\!&\!\! 20 \!\!\!&\!\! --
\!\!\!&\!\!0.352
\!\!\!&\!\!{\bf0.131}
\!\!\!&\!\!0.275
\!\!\!&\!\!0.189
\!\!\!&\!\!0.293
\!\!\!&\!\!0.252
\!\!\!&\!\!\textcolor{blue}{0.249}\!\!\!
\\
& \!\!\! {\small Fast-RCNN (2 fts)~\cite{FastRCNN}} \!\!\!&\!\! 20 \!\!\!&\!\! --
\!\!\!&\!\!0.176
\!\!\!&\!\!0.132
\!\!\!&\!\!0.191
\!\!\!&\!\!0.171
\!\!\!&\!\!0.231
\!\!\!&\!\!{\bf0.189}
\!\!\!&\!\!\textcolor{blue}{0.182}\!\!\!
\\
& \!\!\! {\small Fast-RCNN (1 ft)~\cite{FastRCNN}} \!\!\!&\!\! 30 \!\!\!&\!\! --
\!\!\!&\!\!0.285
\!\!\!&\!\!0.146
\!\!\!&\!\!0.228
\!\!\!&\!\!0.141
\!\!\!&\!\!0.250
\!\!\!&\!\!0.220
\!\!\!&\!\!\textcolor{blue}{0.212}\!\!\!
\\
& \!\!\! {\small Fast-RCNN (2 fts)~\cite{FastRCNN}} \!\!\!&\!\! 30 \!\!\!&\!\! --
\!\!\!&\!\!0.173
\!\!\!&\!\!0.156
\!\!\!&\!\!0.150
\!\!\!&\!\!{\bf0.137}
\!\!\!&\!\!0.132
\!\!\!&\!\!0.221
\!\!\!&\!\!\textcolor{blue}{0.161}\!\!\!
\\
& \!\!\! {\small Ours} \!\!\!&\!\! \textcolor{red}{\bf 10} \!\!\!&\!\! 14.7
\!\!\!&\!\!{\bf0.144}
\!\!\!&\!\!0.146
\!\!\!&\!\!{\bf0.137}
\!\!\!&\!\!0.145
\!\!\!&\!\!{\bf0.122}
\!\!\!&\!\!0.193
\!\!\!&\!\!\textcolor{blue}{\bf0.148}\!\!\!
\\
\hline
\multirow{7}{*}{\rotatebox[origin=c]{90}{Neck\vspace{-3pt}}} &\!\!\! {\small Fast-RCNN (1 ft)~\cite{FastRCNN}} \!\!\!&\!\! \textcolor{red}{\bf 10} \!\!\!&\!\! --
\!\!\!&\!\!0.251
\!\!\!&\!\!0.333
\!\!\!&\!\!0.310
\!\!\!&\!\!0.248
\!\!\!&\!\!0.267
\!\!\!&\!\!0.242
\!\!\!&\!\!\textcolor{blue}{0.275}\!\!\!
\\
& \!\!\! {\small Fast-RCNN (2 fts)~\cite{FastRCNN}} \!\!\!&\!\! \textcolor{red}{\bf 10} \!\!\!&\!\! --
\!\!\!&\!\!0.317
\!\!\!&\!\!0.335
\!\!\!&\!\!0.307
\!\!\!&\!\!0.362
\!\!\!&\!\!0.271
\!\!\!&\!\!0.259
\!\!\!&\!\!\textcolor{blue}{0.309}\!\!\!
\\
& \!\!\! {\small Fast-RCNN (1 ft)~\cite{FastRCNN}} \!\!\!&\!\! 20 \!\!\!&\!\! --
\!\!\!&\!\!0.255
\!\!\!&\!\!0.359
\!\!\!&\!\!0.241
\!\!\!&\!\!0.281
\!\!\!&\!\!0.268
\!\!\!&\!\!0.235
\!\!\!&\!\!\textcolor{blue}{0.273}\!\!\!
\\
& \!\!\! {\small Fast-RCNN (2 fts)~\cite{FastRCNN}} \!\!\!&\!\! 20 \!\!\!&\!\! --
\!\!\!&\!\!0.260
\!\!\!&\!\!0.289
\!\!\!&\!\!0.304
\!\!\!&\!\!0.297
\!\!\!&\!\!0.255
\!\!\!&\!\!0.237
\!\!\!&\!\!\textcolor{blue}{0.274}\!\!\!
\\
& \!\!\! {\small Fast-RCNN (1 ft)~\cite{FastRCNN}} \!\!\!&\!\! 30 \!\!\!&\!\! --
\!\!\!&\!\!0.288
\!\!\!&\!\!0.324
\!\!\!&\!\!0.247
\!\!\!&\!\!0.262
\!\!\!&\!\!0.210
\!\!\!&\!\!0.220
\!\!\!&\!\!\textcolor{blue}{0.258}\!\!\!
\\
& \!\!\! {\small Fast-RCNN (2 fts)~\cite{FastRCNN}} \!\!\!&\!\! 30 \!\!\!&\!\! --
\!\!\!&\!\!0.201
\!\!\!&\!\!0.276
\!\!\!&\!\!0.281
\!\!\!&\!\!0.254
\!\!\!&\!\!0.220
\!\!\!&\!\!0.229
\!\!\!&\!\!\textcolor{blue}{0.244}\!\!\!
\\
& \!\!\! {\small Ours} \!\!\!&\!\! \textcolor{red}{\bf 10} \!\!\!&\!\! 24.5
\!\!\!&\!\!{\bf0.120}
\!\!\!&\!\!{\bf0.144}
\!\!\!&\!\!{\bf0.178}
\!\!\!&\!\!{\bf0.152}
\!\!\!&\!\!{\bf0.161}
\!\!\!&\!\!{\bf0.161}
\!\!\!&\!\!\textcolor{blue}{\bf0.152}\!\!\!
\\
\hline
\multirow{7}{*}{\rotatebox[origin=c]{90}{Nose/Muzzle/Beek\vspace{-3pt}}} &\!\!\! {\small Fast-RCNN (1 ft)~\cite{FastRCNN}} \!\!\!&\!\! \textcolor{red}{\bf 10} \!\!\!&\!\! --
\!\!\!&\!\!0.446
\!\!\!&\!\!0.389
\!\!\!&\!\!0.301
\!\!\!&\!\!0.326
\!\!\!&\!\!0.385
\!\!\!&\!\!0.328
\!\!\!&\!\!\textcolor{blue}{0.363}\!\!\!
\\
& \!\!\! {\small Fast-RCNN (2 fts)~\cite{FastRCNN}} \!\!\!&\!\! \textcolor{red}{\bf 10} \!\!\!&\!\! --
\!\!\!&\!\!0.447
\!\!\!&\!\!0.433
\!\!\!&\!\!0.313
\!\!\!&\!\!0.391
\!\!\!&\!\!0.338
\!\!\!&\!\!0.350
\!\!\!&\!\!\textcolor{blue}{0.379}\!\!\!
\\
& \!\!\! {\small Fast-RCNN (1 ft)~\cite{FastRCNN}} \!\!\!&\!\! 20 \!\!\!&\!\! --
\!\!\!&\!\!0.425
\!\!\!&\!\!0.372
\!\!\!&\!\!0.260
\!\!\!&\!\!0.303
\!\!\!&\!\!0.334
\!\!\!&\!\!0.279
\!\!\!&\!\!\textcolor{blue}{0.329}\!\!\!
\\
& \!\!\! {\small Fast-RCNN (2 fts)~\cite{FastRCNN}} \!\!\!&\!\! 20 \!\!\!&\!\! --
\!\!\!&\!\!0.419
\!\!\!&\!\!0.351
\!\!\!&\!\!0.289
\!\!\!&\!\!0.249
\!\!\!&\!\!0.296
\!\!\!&\!\!0.293
\!\!\!&\!\!\textcolor{blue}{0.316}\!\!\!
\\
& \!\!\! {\small Fast-RCNN (1 ft)~\cite{FastRCNN}} \!\!\!&\!\! 30 \!\!\!&\!\! --
\!\!\!&\!\!0.462
\!\!\!&\!\!0.336
\!\!\!&\!\!0.242
\!\!\!&\!\!0.260
\!\!\!&\!\!0.247
\!\!\!&\!\!0.257
\!\!\!&\!\!\textcolor{blue}{0.301}\!\!\!
\\
& \!\!\! {\small Fast-RCNN (2 fts)~\cite{FastRCNN}} \!\!\!&\!\! 30 \!\!\!&\!\! --
\!\!\!&\!\!0.430
\!\!\!&\!\!0.338
\!\!\!&\!\!0.239
\!\!\!&\!\!0.219
\!\!\!&\!\!0.271
\!\!\!&\!\!0.285
\!\!\!&\!\!\textcolor{blue}{0.297}\!\!\!
\\
& \!\!\! {\small Ours} \!\!\!&\!\! \textcolor{red}{\bf 10} \!\!\!&\!\! 23.8
\!\!\!&\!\!{\bf0.134}
\!\!\!&\!\!{\bf0.112}
\!\!\!&\!\!{\bf0.182}
\!\!\!&\!\!{\bf0.156}
\!\!\!&\!\!{\bf0.217}
\!\!\!&\!\!{\bf0.181}
\!\!\!&\!\!\textcolor{blue}{\bf0.164}\!\!\!
\\
\hline
\end{tabular}}
\caption{Part localization on the Pascal VOC Part dataset. The third and fourth columns show the number of part annotations and the average number of questions for training.}
\label{tab:VOC}
\vspace{-12pt}
\end{table}

\subsection{Experimental results}

We tested our method on the ILSVRC 2013 DET Animal-Part dataset, the Pascal VOC Part dataset, and the CUB200-2011 dataset. We learned AOGs for parts of the head, the neck, and the nose/muzzle/beak of the six animal categories in the Pascal VOC Part dataset. For the ILSVRC 2013 DET Animal-Part dataset and the CUB200-2011 dataset, we learned an AOG for the head part\textcolor{red}{\footnote[4]{It is the ``forehead'' part for birds in the CUB200-2011 dataset.}} of each category. Because the head is shared by all categories in the two datasets, we selected the head as the target part to enable a fair comparison. We did not train the human annotators. During the active QA process, boundaries between two part templates were often very vague, so an annotator could assign a part with either part templates.

\begin{figure*}[t]
\centering
\includegraphics[width=0.9\linewidth]{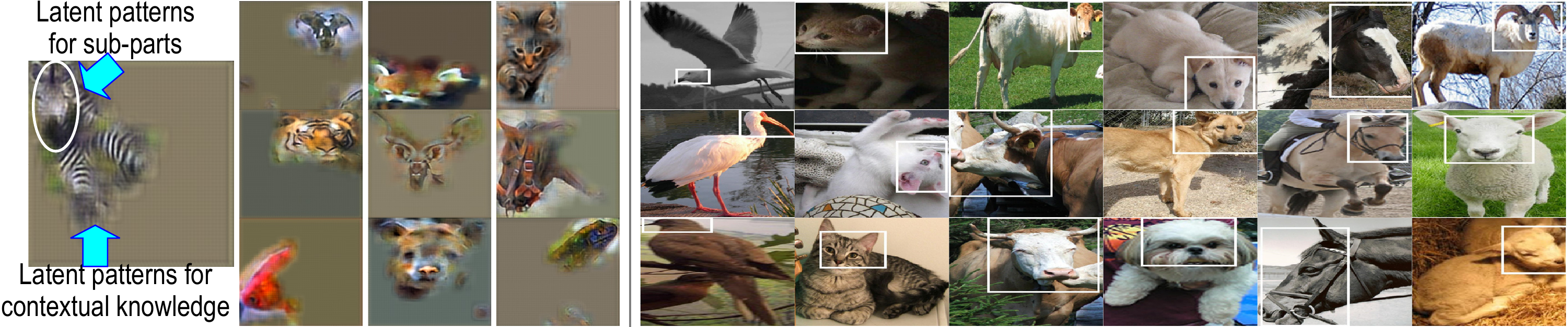}
\caption{Visualization of latent patterns in AOGs for the head part (left) and part localization results based on AOGs (right).}
\label{fig:results}
\vspace{-12pt}
\end{figure*}

In Table~\ref{tab:stat}, we illustrated how the AOG grew when people annotated more parts during the question-answering process. We computed the average number of children for each node in different AOG layers based on the AOGs learned from the PASCAL VOC Part Dataset. It shows that the AOG mainly grew itself by adding new AOG branches for new part templates. The refinement of an AOG branch for an existing part template did not significantly change the size of this AOG branch.

\begin{figure}[t]
\centering
\includegraphics[width=0.99\linewidth]{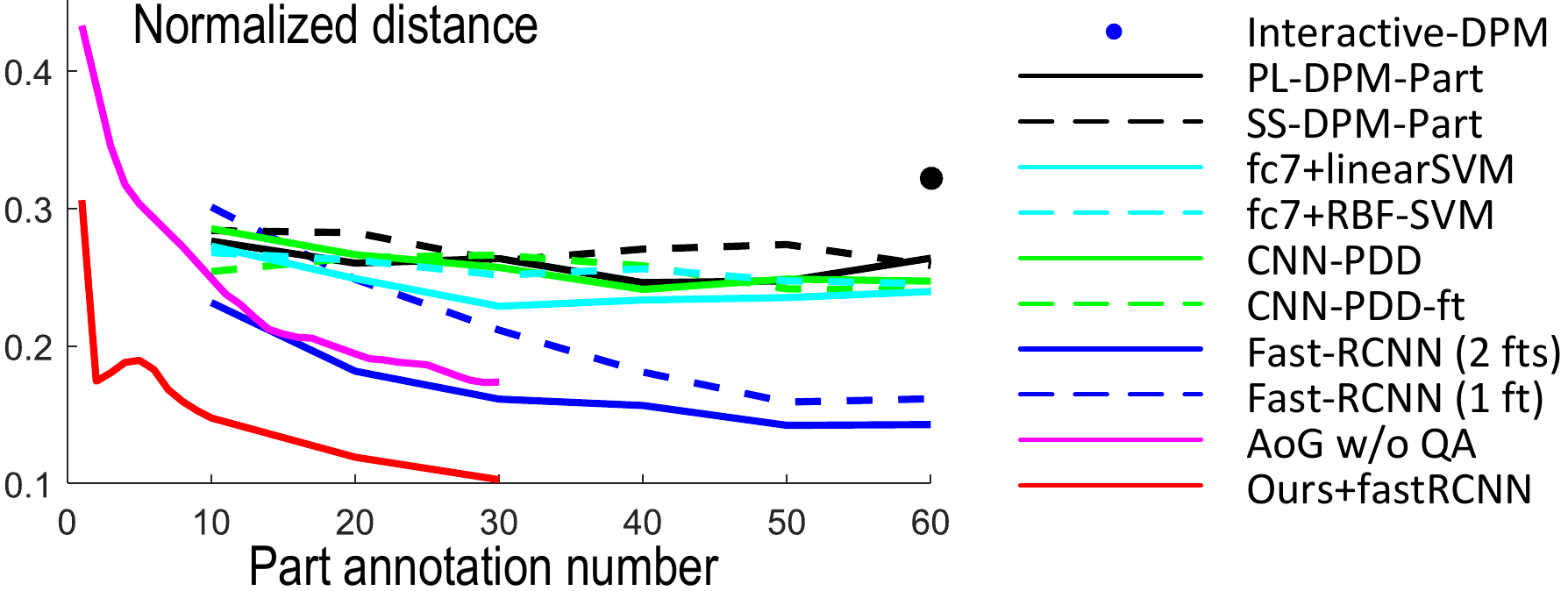}
\caption{Part localization performance on the Pascal VOC Part dataset.}
\label{fig:curve}
\vspace{-12pt}
\end{figure}

Fig.~\ref{fig:results} shows the part localization results based on AOGs and visualizes the content of latent patterns in the AOG based on the technique of \cite{FeaVisual}. Tables~\ref{tab:imgnet}, \ref{tab:VOC}, and \ref{tab:cub200} compares part-localization performance of different baselines on the ILSVRC 2013 DET Animal-Part dataset, the Pascal VOC Part dataset, and the CUB200-2011 dataset, respectively. Tables~\ref{tab:VOC}, and \ref{tab:cub200} show both the number of part annotations and the number of questions. Fig.~\ref{fig:curve} shows the performance of localizing the head part on the PASCAL VOC Part Dataset, when people annotated different number of parts for training. Table~\ref{tab:pcp} shows the results evaluated by the PCP. In particular, the method of \textit{Ours+fastRCNN} combined our method and the fast-RCNN to refine part-localization results\footnote[5]{We used part boxes annotated during the QA process to learn a fast-RCNN for part detection. Given the inference result $\hat{\Lambda}_{V^{\textrm{tmp}}}$ of part template $V^{\textrm{tmp}}$ on image $I$, we define a new inference score for localization refinement $S_{I}^{\textrm{new}}(V^{\textrm{tmp}}\!|\Lambda^{\textrm{new}}_{V^{\textrm{tmp}}})\!=\!S_{I}^{\textrm{old}}(V^{\textrm{tmp}}\!|\hat{\Lambda}_{V^{\textrm{tmp}}})+\lambda_1\Phi(\Lambda^{\textrm{new}}_{V^{\textrm{tmp}}})+\lambda_2\frac{\Vert{\bf p}(\hat{\Lambda}_{V^{\textrm{tmp}}})-{\bf p}(\Lambda^{\textrm{new}}_{V^{\textrm{tmp}}})\Vert}{2\sigma^2}$, where $\sigma=70$ pixels, $\lambda_1\!=\!5$, and $\lambda_2\!=\!10$. $\Phi(\Lambda^{\textrm{new}}_{V^{\textrm{tmp}}})$ denotes the fast-RCNN's detection score for the patch of $\Lambda^{\textrm{new}}_{V^{\textrm{tmp}}}$. ${\bf p}(\Lambda)$ denotes the position of $\Lambda$.}. Our method worked with about $1/6$--$1/2$ part annotations, but exhibited superior performance.

\section{Justification of the methodology}

There are three reasons for the superior performance of our method. First, \textbf{richer information}: the latent patterns in the AOG were pre-fine-tuned using a large number of object images in the category, instead of being learned from a few part annotations. Thus, the knowledge contained in these patterns was far beyond that in the objects with part annotations.

Second, \textbf{less model drift:} Instead of learning/fine-tuning new CNN parameters, our method just used limited part annotations to mine ``reliable'' patterns and organize their spatial relationships to represent the part concept. In addition, during active QA, the computer usually selected and asked about objects with common object poses based on Eq.~(\ref{eqn:predict}), \emph{i.e.} objects sharing some common latent patterns with many other objects. Thus, the learned AOG suffered less from the over-fitting/model-drift problem.

\begin{table}[t]
\centering
\resizebox{0.9\linewidth}{!}{\begin{tabular}{lcc}
\hline
& \# of part annotations & Performance\\
SS-DPM-Part~\cite{SSDPM} & 60 & 7.2\\
PL-DPM-Part~\cite{PLDPM} & 60 & 6.7\\
Part-Graph~\cite{SemanticPart} & 60 & 11.0\\
fc7+linearSVM & 60 & 13.5\\
fc7+RBF-SVM & 60 & 9.5\\
VAE+linearSVM~\cite{VAE} & 30 & 6.7\\
CoopNet+linearSVM~\cite{CoopNet} & 30 & 5.6\\
Fast-RCNN (1 ft)~\cite{FastRCNN} & 30 & 34.5\\
Fast-RCNN (2 fts)~\cite{FastRCNN} & 30 & 45.7\\
Ours+fastRCNN & \textcolor{red}{\bf 10} & 33.0\\
Ours+fastRCNN & \textcolor{red}{\bf 20} & {\bf47.2}\\
Ours+fastRCNN & \textcolor{red}{\bf 30} & {\bf50.5}\\
\hline
\end{tabular}}
\vspace{1pt}
\caption{Part localization performance evaluated using the PCP on the Pascal VOC Part dataset.}
\label{tab:pcp}
\vspace{-12pt}
\end{table}

Third, \textbf{high QA efficiency:} Our QA process balanced both the commonness of a part template and the modeling quality of this part template in Eq.~(\ref{eqn:predict}). In early steps of QA, the computer was prone to asking new part templates, because objects with un-modeled part appearance usually had low inference scores. In later QA steps, common part appearance had been  asked and modeled, and the computer gradually changed to ask about objects of existing part templates to refine certain AOG branches. In this way, our method did not waste much computation in labeling objects that had been well explained or objects with infrequent appearance.

\section{Summary and discussion}

In this paper, we aim to pursue answers to the following three questions: 1) whether we can represent a pre-trained CNN using an interpretable AOG model, which reveals semantic hierarchy of objects hidden in the CNN, 2) whether the representation of the CNN knowledge can be clear enough to let people directly communicate with middle-level AOG nodes, and 3) whether we can let the computer directly learn from weak supervision of active QA, instead of strongly supervised end-to-end learning.

We tested the proposed method for a total of 37 categories in three benchmark datasets, and our method exhibited superior performance to other baselines in terms of weakly-supervised part localization. \emph{E.g.} our method with 11 part annotations performed better than fast-RCNN with 60 part annotations on the ILSVRC dataset in Fig.~\ref{fig:curve}.

\section*{Acknowledgement}
This work is supported by MURI project N00014-16-1-2007 and DARPA SIMPLEX project N66001-15-C-4035.

{\small
\bibliographystyle{ieee}
\bibliography{TheBib}

\begin{thebibliography}{10}\itemsep=-1pt

\bibitem{CNNVisualization_5}
M.~Aubry and B.~C. Russell.
\newblock Understanding deep features with computer-generated imagery.
\newblock {\em In {ICCV}}, 2015.

\bibitem{SSDPM}
H.~Azizpour and I.~Laptev.
\newblock Object detection using strongly-supervised deformable part models.
\newblock {\em In {ECCV}}, 2012.

\bibitem{InteractiveCoseg}
D.~Batra, A.~Kowdle, D.~Parikh, J.~Luo, and T.~Chen.
\newblock Interactively co-segmenting topically related images with intelligent
  scribble guidance.
\newblock {\em In {IJCV}}, 2011.

\bibitem{ActivePart}
S.~Branson, P.~Perona, and S.~Belongie.
\newblock Strong supervision from weak annotation: Interactive training of
  deformable part models.
\newblock {\em In {ICCV}}, 2011.

\bibitem{Gpt_WeaklyCNN}
X.~Chen and A.~Gupta.
\newblock Webly supervised learning of convolutional networks.
\newblock {\em In {ICCV}}, 2015.

\bibitem{SemanticPart}
X.~Chen, R.~Mottaghi, X.~Liu, S.~Fidler, R.~Urtasun, and A.~Yuille.
\newblock Detect what you can: Detecting and representing objects using
  holistic models and body parts.
\newblock {\em In {CVPR}}, 2014.

\bibitem{ChoDiscovery}
M.~Cho, S.~Kwak, C.~Schmid, and J.~Ponce.
\newblock Unsupervised object discovery and localization in the wild:
  Part-based matching with bottom-up region proposals.
\newblock {\em In {CVPR}}, 2015.

\bibitem{OnlineMetric}
Y.~Cong, J.~Liu, J.~Yuan, and J.~Luo.
\newblock Self-supervised online metric learning with low rank constraint for
  scene categorization.
\newblock {\em In {IEEE Transactions on Image Processing}}, 22(8):3179--3191,
  2013.

\bibitem{KB_Fei_Annotation}
J.~Deng, O.~Russakovsky, J.~Krause, M.~Bernstein, A.~Berg, and L.~Fei-Fei.
\newblock Scalable multi-label annotation.
\newblock {\em In {CHI}}, 2014.

\bibitem{FeaVisual}
A.~Dosovitskiy and T.~Brox.
\newblock Inverting visual representations with convolutional networks.
\newblock {\em In {CVPR}}, 2016.

\bibitem{UnsuperCNN}
A.~Dosovitskiy, J.~T. Springenberg, M.~Riedmiller, and T.~Brox.
\newblock Discriminative unsupervised feature learning with convolutional
  neural networks.
\newblock {\em In {NIPS}}, 2014.

\bibitem{Language2VideoAlign}
L.~Duan, D.~Xu, I.~Tsang, and J.~Luo.
\newblock Visual event recognition in videos by learning from web data.
\newblock {\em In {CVPR}}, 2010.

\bibitem{PascalVOC}
M.~Everingham, L.~Gool, C.~Williams, J.~Winn, and A.~Zisserman.
\newblock {\em The PASCAL Visual Object Classes Challenge 2007 (VOC2007)
  Results}.

\bibitem{FastRCNN}
R.~Girshick.
\newblock Fast r-cnn.
\newblock {\em In {ICCV}}, 2015.

\bibitem{VAE}
D.~P. Kingma and M.~Welling.
\newblock Auto-encoding variational bayes.
\newblock {\em In {ICLR}}, 2014.

\bibitem{CNNImageNet}
A.~Krizhevsky, I.~Sutskever, and G.~Hinton.
\newblock Imagenet classification with deep convolutional neural networks.
\newblock {\em In {NIPS}}, 2012.

\bibitem{CNN}
Y.~LeCun, L.~Bottou, Y.~Bengio, and P.~Haffner.
\newblock Gradient-based learning applied to document recognition.
\newblock {\em In {Proceedings of the IEEE}}, 1998.

\bibitem{PLDPM}
B.~Li, W.~Hu, T.~Wu, and S.-C. Zhu.
\newblock Modeling occlusion by discriminative and-or structures.
\newblock {\em In {ICCV}}, 2013.

\bibitem{fineGrained3}
D.~Lin, X.~Shen, C.~Lu, and J.~Jia.
\newblock Deep lac: Deep localization, alignment and classification for
  fine-grained recognition.
\newblock {\em In {CVPR}}, 2015.

\bibitem{MSCOCO}
T.-Y. Lin, M.~Maire, S.~Belongie, L.~Bourdev, R.~Girshick, J.~Hays, P.~Perona,
  D.~Ramanan, C.~L. Zitnick, and P.~Dollar.
\newblock Microsoft coco: Common objects in context.
\newblock {\em In {arXiv:1405.0312v3 [cs.CV]}}, 21 Feb 2015.

\bibitem{CNNFeatureMining}
L.~Liu, C.~Shen, and A.~van~den Hengel.
\newblock The treasure beneath convolutional layers: Cross-convolutional-layer
  pooling for image classification.
\newblock {\em In {CVPR}}, 2015.

\bibitem{Active2}
C.~Long and G.~Hua.
\newblock Multi-class multi-annotator active learning with robust gaussian
  process for visual recognition.
\newblock {\em In {ICCV}}, 2015.

\bibitem{CNNVisualization_2}
A.~Mahendran and A.~Vedaldi.
\newblock Understanding deep image representations by inverting them.
\newblock {\em In {CVPR}}, 2015.

\bibitem{ObjectDiscoveryCNN_1}
M.~Oquab, L.~Bottou, I.~Laptev, and J.~Sivic.
\newblock Is object localization for free? weakly-supervised learning with
  convolutional neural networks.
\newblock {\em In {CVPR}}, 2015.

\bibitem{ObjectDiscoveryCNN_3}
D.~Pathak, P.~Kr\"{a}henb\"{u}hl, and T.~Darrell.
\newblock Constrained convolutional neural networks for weakly supervised
  segmentation.
\newblock {\em In {ICCV}}, 2015.

\bibitem{ImageNet}
O.~Russakovsky, J.~Deng, H.~Su, J.~Krause, S.~Satheesh, S.~Ma, Z.~Huang,
  A.~Karpathy, A.~Khosla, M.~Bernstein, A.~C. Berg, and L.~Fei-Fei.
\newblock Imagenet large scale visual recognition challenge.
\newblock {\em In {IJCV}}, 115(3):211--252, 2015.

\bibitem{KB_Fei_InteractionLabel}
O.~Russakovsky, L.-J. Li, and L.~Fei-Fei.
\newblock Best of both worlds: human-machine collaboration for object
  annotation.
\newblock {\em In {CVPR}}, 2015.

\bibitem{fineGrained2}
K.~J. Shih, A.~Mallya, S.~Singh, and D.~Hoiem.
\newblock Part localization using multi-proposal consensus for fine-grained
  categorization.
\newblock {\em In {BMVC}}, 2015.

\bibitem{MiningAOG}
Z.~Si and S.-C. Zhu.
\newblock Learning and-or templates for object recognition and detection.
\newblock {\em In {PAMI}}, 2013.

\bibitem{ObjectDiscoveryCNN_2}
M.~Simon and E.~Rodner.
\newblock Neural activation constellations: Unsupervised part model discovery
  with convolutional networks.
\newblock {\em In {ICCV}}, 2015.

\bibitem{DiscoveryCNNFeature}
M.~Simon and E.~Rodner.
\newblock Neural activation constellations: Unsupervised part model discovery
  with convolutional networks.
\newblock {\em In {ICCV}}, 2015.

\bibitem{CNNSemanticPart}
M.~Simon, E.~Rodner, and J.~Denzler.
\newblock Part detector discovery in deep convolutional neural networks.
\newblock {\em In {ACCV}}, 2014.

\bibitem{CNNVisualization_3}
K.~Simonyan, A.~Vedaldi, and A.~Zisserman.
\newblock Deep inside convolutional networks: Visualising image classification
  models and saliency maps.
\newblock {\em In {arXiv:1312.6034v2}}, 2013.

\bibitem{VGG}
K.~Simonyan and A.~Zisserman.
\newblock Very deep convolutional networks for large-scale image recognition.
\newblock {\em In {ICLR}}, 2015.

\bibitem{MiddleLevel}
S.~Singh, A.~Gupta, and A.~A. Efros.
\newblock Unsupervised discovery of mid-level discriminative patches.
\newblock {\em In {ECCV}}, 2012.

\bibitem{WeaklyMIL}
H.~O. Song, R.~Girshick, S.~Jegelka, J.~Mairal, Z.~Harchaoui, and T.~Darrell.
\newblock On learning to localize objects with minimal supervision.
\newblock {\em In {ICML}}, 2014.

\bibitem{Language2ActionAlign}
Y.~C. Song, I.~Naim, A.~A. Mamun, K.~Kulkarni, P.~Singla, J.~Luo, D.~Gildea,
  and H.~Kautz.
\newblock Unsupervised alignment of actions in video with text descriptions.
\newblock {\em In {IJCAI}}, 2016.

\bibitem{Active4}
Q.~Sun, A.~Laddha, and D.~Batra.
\newblock Active learning for structured probabilistic models with histogram
  approximation.
\newblock {\em In {CVPR}}, 2015.

\bibitem{TuQA}
K.~Tu, M.~Meng, M.~W. Lee, T.~E. Choe, and S.-C. Zhu.
\newblock Joint video and text parsing for understanding events and answering
  queries.
\newblock {\em In {IEEE MultiMedia}}, 2014.

\bibitem{SelectiveSearch}
J.~R.~R. Uijlings, K.~E.~A. van~de Sande, T.~Gevers, and A.~W.~M. Smeulders.
\newblock Selective search for object recognition.
\newblock {\em In {IJCV}}, 104(2):154--171, 2013.

\bibitem{i13}
S.~Vijayanarasimhan and K.~Grauman.
\newblock Large-scale live active learning: Training object detectors with
  crawled data and crowds.
\newblock {\em {In} CVPR}, 2011.

\bibitem{CUB200}
C.~Wah, S.~Branson, P.~Welinder, P.~Perona, and S.~Belongie.
\newblock The caltech-ucsd birds-200-2011 dataset.
\newblock Technical Report CNS-TR-2011-001, In {California Institute of
  Technology}, 2011.

\bibitem{CoopNet}
J.~Xie, Y.~Lu, S.-C. Zhu, and Y.~N. Wu.
\newblock Cooperative training of descriptor and generator networks.
\newblock {\em In {arXiv 1609.09408}}, 2016.

\bibitem{CNNVisualization_1}
M.~D. Zeiler and R.~Fergus.
\newblock Visualizing and understanding convolutional networks.
\newblock {\em In {ECCV}}, 2014.

\bibitem{fineGrained1}
N.~Zhang, J.~Donahue, R.~Girshick, and T.~Darrell.
\newblock Part-based r-cnns for fine-grained category detection.
\newblock {\em In {ECCV}}, 2014.

\bibitem{CNNAoG}
Q.~Zhang, R.~Cao, Y.~N. Wu, and S.-C. Zhu.
\newblock Growing interpretable graphs on convnets via multi-shot learning.
\newblock {\em In {AAAI}}, 2016.

\bibitem{OurICCV15AoG}
Q.~Zhang, Y.-N. Wu, and S.-C. Zhu.
\newblock Mining and-or graphs for graph matching and object discovery.
\newblock {\em In {ICCV}}, 2015.

\bibitem{CNNSemanticDeep}
B.~Zhou, A.~Khosla, A.~Lapedriza, A.~Oliva, and A.~Torralba.
\newblock Object detectors emerge in deep scene cnns.
\newblock {\em In {ICRL}}, 2015.

\bibitem{CNNSemanticDeep2}
B.~Zhou, A.~Khosla, A.~Lapedriza, A.~Oliva, and A.~Torralba.
\newblock Learning deep features for discriminative localization.
\newblock {\em In {CVPR}}, 2016.

\bibitem{MumfordAOG}
S.~Zhu and D.~Mumford.
\newblock A stochastic grammar of images.
\newblock {\em In {Foundations and Trends in Computer Graphics and Vision}},
  2(4):259--362, 2006.

\end{thebibliography}
}

\newpage

\begin{figure*}[h]
\centering
\includegraphics[width=0.99\linewidth]{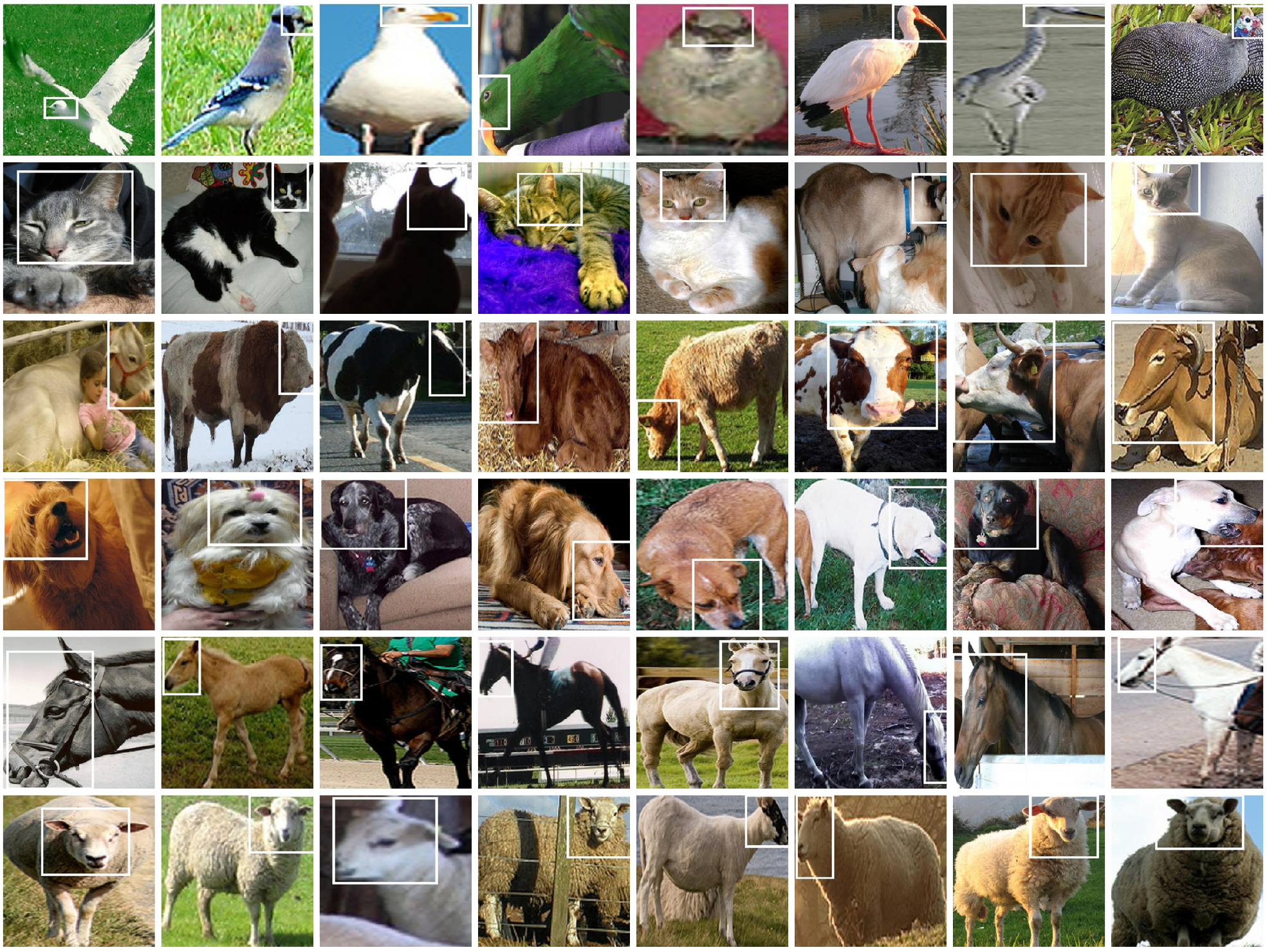}
\caption{Localization results of the head part on animal categories in the Pascal VOC Part dataset~\cite{SemanticPart}}
\end{figure*}

\vspace{15pt}

\begin{figure*}[h]
\centering
\includegraphics[width=0.99\linewidth]{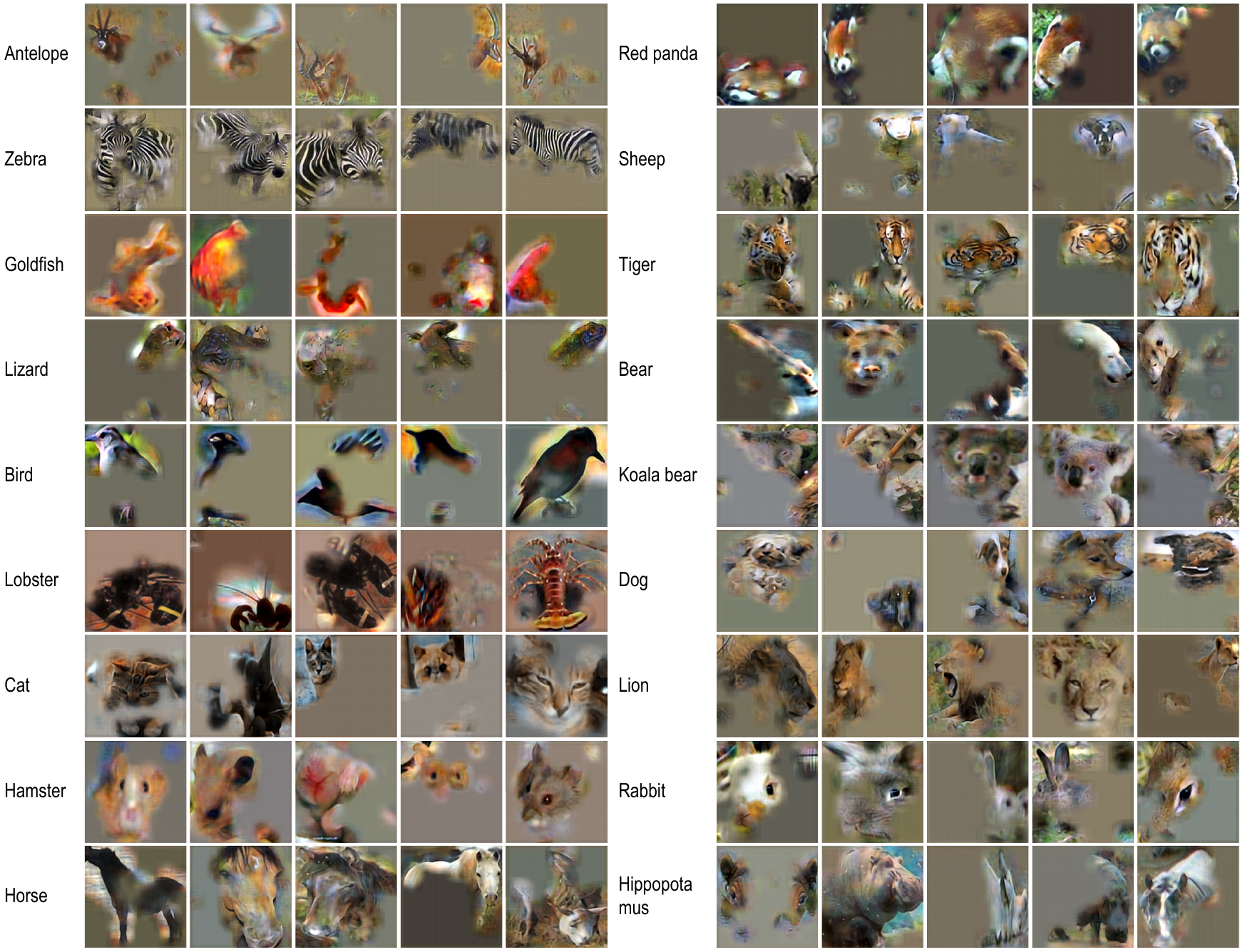}
\caption{Image reconstruction based on AOGs for the head part. In this figure, we only visualize latent patterns located in conv-layers 5--7 based on reconstruction technique of \cite{FeaVisual}. We use neural responses of CNN units in the AOG, which are selected during part parsing, to reconstruct the head part. Some latent patterns in the AOG select CNN units corresponding to constituent regions of the part, while CNN units of other latent patterns represent contexts \emph{w.r.t.} the part.}
\end{figure*}

\end{document}